\documentclass{article}

\usepackage[preprint]{neurips_2026}

\usepackage[utf8]{inputenc}
\usepackage[T1]{fontenc}
\usepackage{microtype}
\usepackage{graphicx}
\usepackage{subcaption}
\usepackage{booktabs}
\usepackage{hyperref}
\usepackage{url}
\usepackage{nicefrac}
\usepackage{xcolor}
\usepackage{amsmath}
\usepackage{amsfonts}
\usepackage{amssymb}
\usepackage{mathtools}
\usepackage{amsthm}
\usepackage{xspace}
\usepackage{kotex}
\usepackage{tabularx}
\usepackage{array}
\usepackage[capitalize,noabbrev]{cleveref}
\usepackage{float}
\usepackage{placeins}
\usepackage[most]{tcolorbox}
\usepackage{listings}
\tcbuselibrary{listings,breakable}
\usepackage{mdframed}
\usepackage{arydshln}

\lstdefinestyle{promptstyle}{
  basicstyle=\ttfamily\small,
  breaklines=true,
  breakatwhitespace=false,
  columns=fullflexible,
  keepspaces=true,
  showstringspaces=false
}

\newtcblisting{promptbox}{
  listing only,
  breakable,
  colback=gray!3,
  colframe=gray!60,
  boxrule=0.5pt,
  arc=2pt,
  left=6pt,right=6pt,top=6pt,bottom=6pt,
  listing options={style=promptstyle}
}

\theoremstyle{plain}

\theoremstyle{definition}

\theoremstyle{remark}

\newcommand{\ToolMATH}{\textsc{ToolMATH}\xspace}
\newcommand{\ToolMATHHard}{\textsc{ToolMATH-Hard}\xspace}

\newif\ifsectionsummary
\sectionsummaryfalse
\newcommand{\SectionSummary}[1]{%
  \ifsectionsummary
    \begin{mdframed}[linewidth=0.5pt]
    #1
    \end{mdframed}
  \fi
}

\title{\ToolMATH: A Diagnostic Benchmark for Long-Horizon Tool Use under Systematic Tool-Catalog Constraints}

\author{%
  Hyeonje Choi \\
  Seoul National University \\
  \texttt{tikiliki0417@snu.ac.kr} \\
  \And
  Jeongsoo Lee \\
  Seoul National University \\
  \texttt{jeongsoo.lee@snu.ac.kr} \\
  \And
  Hyojun Lee \\
  Seoul National University \\
  \texttt{gywnszz@snu.ac.kr} \\
  \And
  Jay-Yoon Lee \\
  Seoul National University \\
  \texttt{lee.jayyoon@snu.ac.kr} \\
}
\begin{document}

\maketitle

\begin{abstract}

We introduce \ToolMATH, a math-grounded diagnostic benchmark for evaluating long-horizon tool use under controllable tool-catalog conditions. \ToolMATH converts stepwise MATH solutions into reusable Python tools with natural-language descriptions and typed schemas, and pairs each problem with a tool environment requiring sequential tool use, intermediate-output reuse, and logically connected tool-call chains. \ToolMATH controls tool availability and catalog difficulty by constructing gold tools and graded distractors with varying similarity to gold tools. \ToolMATH also incorporates behavior-conditioned metrics, enabling diagnostic evaluation beyond final accuracy. Building on these measurements, \ToolMATH emphasizes three evaluation axes: (1) \emph{Adaptability} measures how much Gold-only success is retained when gold tools are replaced entirely by distractors; (2) \emph{Robustness} measures stability under adding distractors as a noise; and (3) \emph{Tool Connectivity} measures whether models preserve accuracy over long executed tool-call chains. Furthermore, trace-level failure analyses characterize how models fail under each tool-catalog condition. Together, these diagnostics reveal distinct model profiles: reliable tool use, tool avoidance, adaptive substitution, and impacts of unreliable tool catalogs. Overall, \ToolMATH provides a controlled testbed for evaluating how language models adapt to changing tool availability, remain robust to distractors, and maintain correctness across long-horizon tool-use trajectories.

\end{abstract}


\SectionSummary{
핵심 논리
\begin{itemize}
    \item 기존 tool use 벤치마크는 (1) multi-tool이라는 이름과는 달리 tool call 횟수가 적거나, (2) tool들 간 logical connectivity가 약하거나, (3) 비슷하거나 동일한 tool들이 혼재하거나(redundancy), (4) 필요한 tool이 없는(insufficiency) 실제 tool의 환경의 난점을 회피하는 경우가 많다.
    \item 실제 환경에서 진짜 어려운 것은 multi-hop의 tool sequence와 유사하면서도 혼동을 유발할 수 있는 tool들 속에서의 선택, 그리고 필요한 tool이 없는 경우에 대한 fail-safe이다(tool을 고르지 않고 알아서 답하거나, 답하지 못한다고 말하거나).
    \item 즉 intro에서는 왜 ToolMATH가 필요한가를 one-liner로 압축해서 설명하고, 그 필요성의 근거를 위의 근거들을 발전시킨 후 연결한 뒤, ToolMATH가 이 세 축을 어떻게 측정하는지(tool extraction, validation, distractor/Distractors-only setting evaluation)를 정리함.
\end{itemize}
TODO
\begin{itemize}
    \item ToolMATH가 domain을 수학으로 선택한 이유를 한 단락으로: 정답 채점이 명확하고(correctness), 해법이 논리적이고 단계적으로 분해되며(logical connectivity), 여러 solution step 간 작은 오류조차 오답으로 (robustness check).
    \item 대표 문제 1개(question + tool list + tool call sequence)를 figure로 미리 보여주기.
    \item 최종 데이터 규모(최종 validation 후 총 tool number, question number) 숫자 확정.
\end{itemize}
}

\section{Introduction}
\label{sec:intro}

Tool-augmented language models are increasingly used as interfaces to APIs, code, and structured utilities \citep{schick2023toolformer,qin2023toollm,patil2023gorilla,li2023apibank}.
In realistic deployments, however, models rarely operate with a small and perfectly specified tool set.
Instead, they often face large tool catalogs assembled by retrieval or system design, where multiple tools appear relevant, distractor tools overlap semantically with intended tools, and some required capabilities may be missing \citep{karpas2022mrkl,patil2025bfcl}.
Evaluating tool-augmented models therefore requires more than measuring final answer accuracy under a fixed tool list: we need to measure how models adapt when intended tools disappear, how robustly they behave as distractors become more similar, and whether they can maintain correctness across logically connected multi-step tool-call chains.

We introduce \ToolMATH, a math-grounded diagnostic benchmark for controlled evaluation of long-horizon tool use under systematically varied tool-catalog conditions.
\ToolMATH converts human-annotated MATH solution steps into reusable Python tools with natural-language descriptions and typed schemas, and pairs each problem with a tool environment that requires sequential tool use, intermediate-output reuse, and multi-step tool composition \citep{hendrycks2021math}.
Since tools are exposed only through names, descriptions, and input schemas, while implementations remain hidden, \ToolMATH directly evaluates tool selection, parameterization, and observation-conditioned reasoning rather than direct access to code. The benchmark comprises approximately 8k questions and 12k tools, alongside a curated hard split, \ToolMATHHard, containing 329 challenging questions and 362 tools.

We choose mathematics as the construction domain for three reasons.
First, mathematical answers enable objective automatic scoring.
Second, stepwise solutions naturally induce logically connected tool-use trajectories, allowing us to evaluate whether models can compose tool calls over multiple dependent steps.
Third, many problems admit alternative solution paths, which makes it possible to observe whether models can still solve some instances when gold tools are removed and only distractor tools remain.

\paragraph{Validation.}
To make large-scale evaluation reliable, \ToolMATH applies a sequential validation-and-repair pipeline.
Tool-wise validation checks description--implementation consistency through test-case executions, while question-wise validation retains episodes that are empirically solvable with the translated Python tools under the intended tool-use protocol.
Ambiguous or inconsistent cases undergo targeted human audit and repair, followed by re-validation.

\paragraph{Controlled tool-catalog conditions.}
A central design goal of \ToolMATH is controllability.
For each problem, we construct a gold tool set and sample distractor tools under graded similarity levels to the gold tools.
This yields three evaluation conditions: the \emph{Gold-only condition}, where only intended tools are available; the \emph{Gold-present condition}, where gold tools are mixed with distractors; and the \emph{Distractors-only condition}, where gold tools are removed and only distractors remain.
These conditions allow us to separate three deployment-facing axes: long-horizon tool connectivity, distractor robustness, and missing-capability adaptation.

\paragraph{Diagnostic metrics.}
Beyond final accuracy, \ToolMATH supports behavior-conditioned and environment-sensitive diagnostics.
Tool-call rate, Tool-Acc, and NoTool-Acc characterize whether models solve by using tools, avoiding tools, or over-calling tools in unreliable catalogs.
In addition, we introduce metrics that summarize how model behavior changes as the tool environment changes.
\emph{Adaptability} measures how much Gold-only success is retained when gold tools are entirely replaced by distractors.
\emph{Robustness} measures how stable this retention remains across distractor similarity levels.
\emph{Tool connectivity} evaluates whether models preserve accuracy as the number of executed tool calls increases, capturing their ability to connect intermediate tool outputs over longer trajectories.

In summary, our contributions are:
(i) a math-grounded benchmark of reusable hidden-code tools and long-horizon tool-composition episodes derived from MATH solutions;
(ii) a validated tool-construction pipeline that reduces benchmark noise at scale;
(iii) controlled Gold-only, Gold-present, and Distractors-only conditions with graded distractor similarity for evaluating redundancy, missing capabilities, and catalog difficulty; and
(iv) diagnostic metrics and trace-level analyses for measuring behavior-conditioned accuracy, which are Adaptability, Robustness, Tool Connectivity, and 
(v) failure co-occurrence for models with different environments.
Together, these components make \ToolMATH a controlled testbed for studying how language models adapt to changing tool availability, remain robust to distractors, and maintain correctness across long-horizon tool-use trajectories.

\SectionSummary{
핵심 논리
\begin{itemize}
    \item 관련 연구에서 (1) LLM의 tool calling 연구가 어떤 문제의식에서 출발했고, (2) 관련된 방법론 및 벤치마크들이 tool use를 어떤 형태로 정식화했으며, (3) 수학 추론 데이터셋(MATH 등)과 기존 tool-use 벤치마크들이 어떤 평가 축을 제공해왔는지의 흐름을 정리한다.
    \item tool calling 연구의 큰 흐름은 LLM이 tool call을 갖추게 하는 방향(tool 선택, parameter 구성, execution에 대한 feedback)으로 발전해 왔고, 이후에는 robust multi-step tool call 등으로 확장되었음.
    \item 방법론에 대한 내용(ReAct) 정리 : 단일 정답뿐 아니라 실행 과정(중간 단계)까지도 평가할 수 있는 기반 마련
    \item 한편 수학 데이터셋(MATH 등)은 정답 채점이 명확하고 해법이 단계적으로 분해된다는 장점 때문에, 본 논문에서 tool-use/execution based evaluation을 설계하기 위한 바탕으로 활용할 수 있었음. ToolMATH는 특히 다른 문제(query)에서도 사용 가능한 general tool environment와 redundancy/insufficiency를 결합하는 방향으로 벤치마크를 설계한다.
\end{itemize}
}

\section{Related Work}
\label{sec:related}

Prior work on tool-augmented language models spans learning to invoke tools and APIs, as well as benchmark suites that standardize function-calling formats and evaluation protocols \citep{schick2023toolformer,li2023apibank,qin2023toollm,patil2023gorilla,xu2023toolbench,patil2025bfcl}.
Another line of work studies deployment-facing failures under imperfect tool access, including missing capabilities and underspecified tasks \citep{trevino2025failtalms}.
Separately, tool-use methods and agentic interfaces such as ReAct and controller-style systems focus on how models interleave reasoning with actions and orchestrate multi-step interactions \citep{yao2023react,shen2023hugginggpt}.
Finally, structured task generators explicitly model tool dependencies, e.g., via graph-based construction of tool-use tasks \citep{shen2024taskbench}.

Math reasoning datasets such as GSM8K and MATH provide correctness-checkable targets and naturally decomposed multi-step solution structure \citep{cobbe2021training,hendrycks2021math}.
We use MATH's annotated solution as a construction substrate: extracting reusable tool primitives from solution steps and forming long-horizon tool-composition episodes with deterministic scoring.

Overall, existing work addresses important aspects of tool use---function-calling standardization, imperfect tool access, agentic controllers, and structured dependency construction---but typically focuses on one or two axes at a time.
\ToolMATH differs by combining deterministic math-grounded scoring with reusable hidden-code tools, logical-hop annotations, graded gold--distractor similarity, and \emph{Distractors-only} missing-capability settings in a single benchmark.
This combination enables controlled analysis of how sequential dependence, catalog redundancy, and tool insufficiency jointly affect model behavior.

\SectionSummary{
핵심 논리
\begin{itemize}
    \item ToolMATH가 무엇으로 구성되는지(정의)와 왜 그 구성이 평가를 가능하게 만드는지(설계 논리)를 설명함.
    \item 하나의 문제를 풀 때 각 문제에 대응되는 tool set을 정의하고, 각 tool을 (name, description, input schema, python code)로 정의함. 이는 일반적으로 tool call protocol에서 쓰이는 MCP를 본뜬 것임.
    \item Tool Extraction : ToolMATH는 MATH에서의 각 문제의 solution step을 재사용 가능한 Python Function 단위로 분해해 tool로 정의해서, 문제를 풀기 위해 여러 tool을 sequential하게 호출해야 하는 구조를 만든다. 이를 통해 만들어진 multi-tool scenario가 자연스럽게 evaluation의 대상이 된다.
    \item Tool-wise Validation : 가장 큰 evaluation error가 될 수 있는 원인인 tool description-code 간의 불일치를 제거하기 위해 tool-wise validation으로 이 불일치를 최대한 제거함. 모델 성능을 재는 것이 아니라, 벤치마크 자체의 신뢰도를 확보하는 단계임.
    \item Question-wise validation : tool이 충분히 있다고 해도 정작 문제를 풀 때는 의미없는 tool이거나, 실제로 해당 tool을 call한다 해도 풀리지 않는 문제들이 발생하므로, question-wise validation으로 solvable한 question과 최소 1회 이상 call된 tool만을 남긴다.
    \item 마지막으로 deterministic한 level별 distractor와 Distractors-only 구성을 통해, tool이 많고 유사한 환경 및 필요한 tool이 없는 환경을 통제변수로 만들고, 이 변수가 각 모델별, 그리고 방법론 별 failure case에 어떻게 영향을 주는 지 분석한다.
\end{itemize}
TODO
\begin{itemize}
    \item 위 내용을 전부 아우르는 figure 그리기
    \item tool extraction에 대한 구체적인 내용(prompt 등)은 appendix로
\end{itemize}
}

\section{The \ToolMATH Dataset}
\label{sec:dataset}

We describe the tool set and problem instances that underlie \ToolMATH, including how we form the final benchmark from complementary validation signals, and how we annotate each problem's logical-hop complexity for evaluation.

\subsection{Overview}

\paragraph{Dataset and tool definitions.}
We define \ToolMATH as a collection of tool-grounded problem instances, where each instance is a pair $(p,\mathcal{S}(p))$ consisting of a MATH problem $p$ and a tool environment $\mathcal{S}(p)$ exposed to the model. Derived from the MATH dataset \citep{hendrycks2021math}, each tool within this environment is explicitly defined by: (i) a \emph{name} (Python function name), (ii) a \emph{natural-language description} specifying semantics and parameters, (iii) a typed \emph{input schema}, (iv) a deterministic \emph{Python implementation}, and (v) metadata linking it to source problems and solution steps. We extract raw tools from MATH solution steps and validate them to form the final evaluation set, which comprises $12{,}369$ tools and $7{,}699$ questions.

\subsection{Design Objectives}

\ToolMATH is designed around four objectives: 
(i) tools should be reusable mathematical operations rather than problem-specific macros; 
(ii) natural-language descriptions and typed schemas should align with executable behavior; 
(iii) tools should support multi-step solution plans with measurable logical-hop complexity; and 
(iv) distractors should be plausible rather than trivially irrelevant, enabling controlled evaluation of misleadingly similar tools, missing-capability settings, and alternative valid solution trajectories.

\begin{figure}[t]
  \centering
  \includegraphics[width=0.8\textwidth]{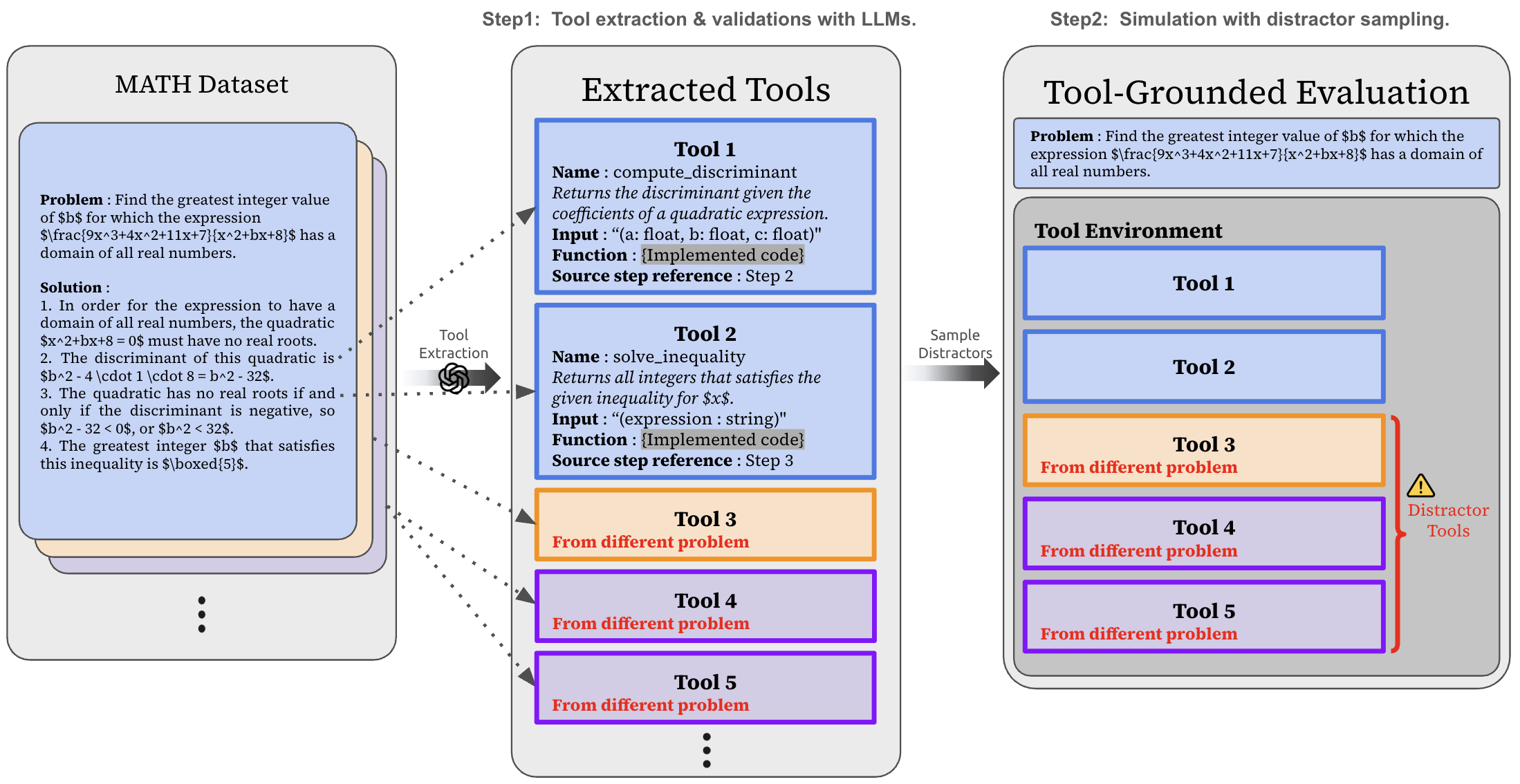}
  \caption{\ToolMATH construction and evaluation pipeline.
  \textbf{Step 1: Tool extraction and validation.} We convert MATH solution steps into schema-specified Python tools and retain only tools whose described semantics are consistent with their executed behavior via validation.
  \textbf{Step 2: Tool-grounded evaluation with controlled redundancy.} For each problem, we form a tool environment by combining its gold tools with distractor tools sampled from other problems.}
  \label{fig:toolmath-pipeline}
\end{figure}

\subsection{Construction Pipeline}
\label{sec:construction}

Figure~\ref{fig:toolmath-pipeline} outlines the pipeline from MATH solutions to tools and tool environments.

\subsubsection{From MATH Solutions to Tools}

We begin from annotated MATH solutions and decompose solution steps into reusable operations.
For each step, we define a tool with (i) a typed signature, (ii) a Python implementation, and (iii) a description that is sufficient for a model to infer correct usage without seeing code.
For extracted tools that are problem-specific or underspecified, we generalize them by abstraction, clarifying constraints and edge cases, and enriching descriptions to reduce ambiguity.

\subsubsection{Validation and Final Dataset Composition}
\label{sec:final-dataset}

\ToolMATH applies a three-stage validation-and-repair pipeline to ensure that benchmark errors are not conflated with model errors.
First, \emph{tool-wise validation} checks whether each tool's natural-language description and typed schema agree with its executed behavior using schema-valid test cases and multi-model judgments.
Second, \emph{question-wise validation} runs the same Plan+ReAct protocol used at evaluation time and retains episodes for which at least one gold tool is used in a correct trace.
Third, ambiguous or rejected cases undergo human audit and repair, followed by re-validation.
Full validator prompts, thresholds, model lists, and audit guidelines are provided in Appendix~\ref{app:toolwise_system_prompt}.

\subsection{\ToolMATHHard}
\label{subsec:hardset}

To provide a more tool-reliant stress setting, we additionally curate \ToolMATHHard.
It contains 329 questions for which automatically extracted gold tools did not yield a validated tool-use trajectory under our trace-based criterion.
For this split, we provide 362 human-authored gold tools and validate them separately, making \ToolMATHHard a stress test for difficult instances in the perspective of automatic tool extraction.
We use \ToolMATHHard for additional analyses of tool availability and framework sensitivity.

\subsection{Logical-Hop Annotation}
\label{subsec:hops}

For each problem $p$, we assign a hop count $h(p)$ based on the extracted solution structure.
Intuitively, $h(p)$ measures how many dependent tool uses are required.
We use $h(p)$ as an evaluation axis to separate failures associated with increased overlap between gold and non-gold tools from failures due to long-horizon planning under tight intermediate dependence (Section~\ref{sec:results}).

\SectionSummary{
핵심 논리
\begin{itemize}
    \item ToolMATH가 어떻게 실행하고 측정하는지에 대한 정의
    \item 평가의 대상이 되는 문제들을 (distractor 난이도, distractor 개수, 문제의 logical hop, Distractors-only 여부)으로 정의하고, 모델의 방법론 별로 출력을 기록함. (즉, 서로 다른 방법론에 따라 서로 다른 failure case를 유형화함.)
    \item 모델이 보는 정보는 tool의 이름과 description, 그리고 input schema임. 구현 코드는 보지 못하기 때문에, 성능은 자연어 기반의 mapping과 parameter filling 능력에 의해 강하게 결정됨.
    \item 평가 조건은 gold-only, gold+distractor, Distractors-only로 나누고, heatmap은 (i) distractor 개수 k--sampling level l, (ii) distractor 개수 k--문제의 logical hop(이때 l=3 고정)으로 제시함. 이때 k가 증가할수록 gold와 유사한 distractor tool이 포함된다는 속성을 의도해 비교의 공정성을 확보한다.
    \item 지표는 정답률로 자동 측정하되, 각 방법론 별로 몇몇 문제를 random sampling해서 failure case에 대한 유형화를 함.
\end{itemize}
}

\section{Experimental Setup}
\label{sec:experiments}

\subsection{Tool-Set Construction and Distractor Settings}
\label{sec:distractors}

We evaluate robustness under both \emph{redundancy} (distractors) and \emph{insufficiency}.
For each problem \(p\), let \(\mathcal{G}(p)\) denote its gold tool set and \(\mathcal{U}\) the global pool of filtered, validated tools. 
For each distractor--gold similarity level \(\ell \in \{1,\dots,5\}\) and distractor budget \(k \in \{5,10,20,50\}\), we sample distractors \(\mathcal{D}_{\ell,k}(p) \subseteq \mathcal{U}\setminus\mathcal{G}(p)\). 
We define three tool-catalog conditions throughout the paper.
In the \textbf{Gold-only condition}, the model is given only the gold tools: \( \mathcal{S}^{\mathrm{gold\text{-}only}}(p)=\mathcal{G}(p). \)
In the \textbf{Gold-present condition}, the model is given the gold tools together with sampled distractor tools: \(\mathcal{S}^{\mathrm{gold\text{-}present}}_{\ell,k}(p)=\mathcal{G}(p)\cup\mathcal{D}_{\ell,k}(p).\), including Gold-only as the special case with \(k=0\). 
The \textbf{Distractors-only condition} removes \(\mathcal{G}(p)\) and exposes only \(\mathcal{D}_{\ell,k}(p)\). Thus, the Gold-present condition evaluates redundancy and overlap in the presence of the intended tools, whereas the Distractors-only condition evaluates missing-capability settings where the intended tools are absent.

\paragraph{Similarity levels.}
We define five distractor sampling levels \(\ell \in \{1,\dots,5\}\), each inducing a different candidate distribution over \(\mathcal{U}\setminus\mathcal{G}(p)\) and thereby controlling how similar distractors are to the gold tools. 
At \textbf{Level 1 (different-category random)}, we sample from tools whose source-problem category differs from \(p\)'s category annotated from the MATH dataset. 
At \textbf{Level 2 (pure random)}, we sample uniformly from the entire pool \(\mathcal{U}\setminus\mathcal{G}(p)\). 
At \textbf{Level 3 (same-category random)}, we sample uniformly from tools from the same category as \(p\). 
At \textbf{Level 4 (embedding similarity retrieval)}, we rank candidates by description-embedding similarity to the gold tools. 
At \textbf{Level 5 (keyword overlap + embedding)}, we rank candidates by keyword overlap with the gold tools and break ties by embedding similarity. 
Higher levels correspond to increasing semantic overlap with the gold tools, shifting from low-overlap random sampling (Levels~1--3) to retrieval-based high-overlap sampling (Levels~4--5).

\paragraph{Nested distractor guarantee.}
For each fixed \((p,\ell)\), we precompute an ordered list of 100 distractors and define \(\mathcal{D}_{\ell,k}(p)\) as its first \(k\) elements.
This enforces \(\mathcal{D}_{\ell,k_1}(p)\subseteq \mathcal{D}_{\ell,k_2}(p)\) whenever \(k_1<k_2\), ensuring comparisons across \(k\) reflect increasing distractor density rather than different samples.

\paragraph{Safeguards and reproducibility.}
We enforce globally unique tool names by appending suffixes when base names collide (e.g., \texttt{solve\_linear\_a}, \texttt{solve\_linear\_b}, \dots).
All sampling is deterministic given a fixed seed and serialized tool pool.
Full implementation details, including embedding model choice, top-100 retrieval, and the exact nested construction, are provided in Appendix~\ref{app:distractor_sampling}.

\subsection{Models}
\label{sec:models}


We evaluate a mix of proprietary and open models: \textbf{GPT-5}, \textbf{Claude-4.6}, \textbf{Gemini-3.1-Pro}, \textbf{GPT-4o-mini}, \textbf{Llama 3-8B}\citep{grattafiori2024llama3}, and \textbf{Qwen 2.5-7B}\citep{yang2024qwen25}.

\subsection{Tool-Use Protocols}
\label{sec:tool-protocols}

\paragraph{Main protocol (Plan+ReAct).}
Unless stated otherwise, we use a planning-augmented ReAct protocol \citep{yao2023react}.
The model first writes a brief plan and then alternates between (i) an intermediate reasoning step and (ii) a structured tool call (tool name + JSON arguments conforming to schemas), conditioned on executed tool outputs.
For each run, we log the full tool-call trace and intermediate text. Detailed prompts for each tool-use protocol are reported in Appendix~\ref{app:protocol_prompts}.

\subsection{Evaluation Conditions and Metrics}
\label{sec:conditions}

We evaluate three tool-catalog conditions: the \textbf{Gold-only condition}, the \textbf{Gold-present condition}, and the \textbf{Distractors-only condition}.
The Gold-only condition provides only the gold tools.
The Gold-present condition provides the gold tools together with \(k \in \{5,10,20,50\}\) distractors sampled at similarity Level~\(\ell \in \{1,\dots,5\}\).
The Distractors-only condition removes the gold tools and provides only the corresponding distractors.
We also report a \textbf{No tools baseline}, in which the model answers without access to any tools; this is a baseline, not a tool-catalog condition.
We report exact-match answer accuracy after standard normalization.
We also analyze traces via tool-call counts and tool identity distributions.

\paragraph{Behavior-conditioned metrics.}
In addition to overall exact-match accuracy, we report three behavior-conditioned metrics that characterize how models use or avoid tools under each catalog condition. Tool-call rate is the fraction of examples with at least one valid tool call. Tool-Acc is accuracy conditioned on executing at least one valid tool call, while NoTool-Acc is accuracy conditioned on producing an answer without any valid tool call. When a conditioning denominator is zero, the corresponding metric is omitted. These metrics separate final task success from tool-use behavior: two models can have similar accuracy while differing substantially in whether they solve directly, use tools selectively, or over-call tools in distractor-heavy environments.

\subsection{Human Error Analysis}
\label{sec:error-analysis}

For trace diagnostics, we sample failed traces and assign non-exclusive failure labels using full reasoning/tool-call trajectories.
The annotation protocol, taxonomy, and marginal failure distributions are reported in Appendix~\ref{sec:results-failurecases}, while co-occurrence patterns are analyzed in Section~\ref{sec:main-failure}.

\SectionSummary{
핵심 논리
\begin{itemize}
    \item ToolMATH가 왜 필요한지를 결과 형태로 보여줌 : 기존 벤치마크에서는 잘 드러나지 않는 failure case가 ToolMATH의 환경에서 체계적으로 드러난다는 점을 강조
    \item 두 가지 heatmap을 제시함: (1) distractor 개수--sampling level, (2) distractor 개수--문제의 logical hop(이때 sampling level은 3으로 고정)
    \item 첫 번째는 distractor에 대한 robustness, 두 번째는 현실적인 tool set에서의 계획 능력(모델의 planning capability)을 진단하는 목적임
\end{itemize}
TODO
\begin{itemize}

\end{itemize}
}

\section{Results and Analysis}
\label{sec:results}

We report results along two difficulty axes: (i) distractor--gold similarity level (Levels~1--5) and distractor density ($k$), and (ii) intrinsic compositional difficulty measured by the number of logical hops.
Unless stated otherwise, accuracies are computed under the main Plan+ReAct protocol described in \cref{sec:tool-protocols}.
We also include a No tools baseline where the model answers without any tool calls.

\paragraph{Gold-present difficulty trends.}
Detailed representative curves are provided in Appendix~\ref{app:gold-present-trends}, and full hop-by-budget heatmaps are provided in Appendix~\ref{sec:appendix-gold}; we summarize the main trend here.
Accuracy decreases with logical-hop count even when gold tools are available, indicating that tool availability alone does not ensure successful long-horizon execution.
High-similarity distractors further amplify this degradation, especially in higher-hop regimes.
Thus, logical-hop complexity is the dominant intrinsic difficulty axis, while distractor similarity acts as an amplifying catalog-level factor.

\subsection{Behavior-conditioned tool-use profiles}
\label{sec:behavior-profiles}

Beyond final accuracy, \ToolMATH allows us to compare how models use tools under different tool-catalog conditions.
\textbf{Accuracy} is exact-match accuracy over all evaluated questions.
\textbf{Tool-Acc} is accuracy restricted to questions where the model executes at least one valid tool call, and \textbf{NoTool-Acc} is accuracy restricted to questions where the model produces an answer without executing any valid tool call.
\textbf{Tool-call rate} is the percentage of questions with at least one valid tool call.
Table~\ref{tab:behavior_metrics} reports behavior-conditioned performance under three tool-catalog conditions.
Unlike a single accuracy score, these metrics separate whether a model succeeds by using tools or by answering independently, and track its tool utilization rates across various conditions.

\begin{table}[t]
\centering
\caption{
\textbf{Gold-only condition} provides only the gold tools.
\textbf{Gold-present condition} provides gold tools together with the same distractor budget of k = 5 with Level 3 sampling.
\textbf{Distractors-only condition} removes the gold tools and provides only the same distractor budget of k = 5 with Level 3 sampling.
All values are percentages.
}
\label{tab:behavior_metrics}
\small
\setlength{\tabcolsep}{3pt}
\begin{tabular}{llrrrr}
\toprule
\textbf{Model} & \textbf{Condition} & \textbf{Accuracy} & \textbf{Tool-Acc} & \textbf{NoTool-Acc} & \textbf{Tool-call rate} \\
\midrule
Claude-4.6 & Gold-only          & 98.4 & 97.9 & 99.5 & 65.6 \\
Claude-4.6 & Gold-present & 97.8 & 96.5 & 99.4 & 67.8 \\
Claude-4.6 & Distractors-only   & 97.8 & 98.1 & 97.8 & 15.2 \\
\hdashline
Gemini-3.1-Pro & Gold-only          & 99.2 & 90.4 & 99.7 & 5.4 \\
Gemini-3.1-Pro & Gold-present & 99.0 & 97.7 & 99.3 & 18.9 \\
Gemini-3.1-Pro & Distractors-only   & 98.9 & 99.4 & 98.8 & 17.6 \\
\hdashline
GPT-5 & Gold-only          & 99.2 & 100.0 & 99.2 & 3.9 \\
GPT-5 & Gold-present & 99.2 & 100.0 & 99.2 & 5.1 \\
GPT-5 & Distractors-only   & 99.0 & 100.0 & 99.0 & 0.3 \\
\hdashline
GPT-4o-mini & Gold-only          & 57.9 & 58.8 & 28.3 & 86.4 \\
GPT-4o-mini & Gold-present & 55.3 & 55.7 & 25.9 & 86.2 \\
GPT-4o-mini & Distractors-only   & 49.8 & 40.8 & 39.7 & 67.2 \\
\hdashline
Llama3-8B & Gold-only          & 21.6 & 23.7 & 0.2 & 90.7 \\
Llama3-8B & Gold-present & 14.6 & 15.8 & 1.4 & 86.4 \\
Llama3-8B & Distractors-only   & 8.1  & 8.7  & 0.2 & 92.9 \\
\hdashline
Qwen2.5-7B & Gold-only          & 59.6 & 60.9 & 31.0 & 89.1 \\
Qwen2.5-7B & Gold-present & 60.6 & 60.5 & 52.0 & 75.6 \\
Qwen2.5-7B & Distractors-only   & 56.3 & 52.1 & 51.2 & 69.0 \\
\bottomrule
\end{tabular}
\end{table}

The results show that \ToolMATH distinguishes models not only by final accuracy, but also by their tool-use policy under different catalog conditions.
Among recent proprietary models, all three maintain high overall accuracy across the Gold-only, Gold-present, and Distractors-only conditions, but they do so through different behaviors.
GPT-5 is highly conservative: it rarely calls tools, especially in the Distractors-only condition, and its observed Tool-Acc remains high when tools are called.
Claude-4.6 is substantially more tool-forward, with high Tool-call rate in the Gold-only and Gold-present conditions, while still maintaining high Tool-Acc.
Gemini-3.1-Pro lies between these regimes, increasing tool use when distractors are present while preserving strong overall accuracy.

Older proprietary and open-source models exhibit a different pattern.
GPT-4o-mini, Llama3-8B, and Qwen2.5-7B call tools more frequently, but their Tool-Acc drops more sharply when the tool catalog becomes less reliable, especially in the Distractors-only.
Llama3-8B is the clearest example: it calls tools in over 90\% of Distractors-only instances, yet its Tool-Acc remains below 10\%.
This indicates that the benchmark exposes not only whether a model solves a task, but whether it can calibrate tool use to tool availability.

\subsection{Adaptability and Robustness under tool-catalog shifts}
\label{sec:adaptability-robustness}

We next examine how model success changes when the tool catalog is systematically altered.
Final accuracy alone does not distinguish whether a model succeeds because it truly uses the intended tools, because it can recover through alternative tool trajectories or internal reasoning, or because it remains stable under additional distractors.
To capture these tendencies, we define a common \emph{performance retention ratio} (PRR).
For each model, distractor similarity level \(\ell\), and target condition \(C\), PRR is the fraction of examples that are answered correctly in the Gold-only condition and remain correct under condition \(C\):
\[
\mathrm{PRR}_{\ell}^{C}
=
\frac{
\left|
\left\{
q:
\mathrm{Correct}_{\mathrm{Gold}}(q)=1
\land
\mathrm{Correct}_{C,\ell}(q)=1
\right\}
\right|
}{
\left|
\left\{
q:
\mathrm{Correct}_{\mathrm{Gold}}(q)=1
\right\}
\right|
}.
\]
Thus, PRR measures the preservation of a model's success after the tool environment changes.

We use this common PRR definition to derive two diagnostics.
First, \emph{Adaptability} (\(\mathrm{PRR}_{1}^{\mathrm{DO}}\)) measures the retention of model performance when gold tools are replaced entirely by distractors (i.e., the Level~1 Distractors-only condition). Higher Adaptability indicates the model can compensate for missing gold tools via internal reasoning or alternative tool paths.
Second, level-wise \emph{Robustness} (\(\mathrm{PRR}_{\ell}^{\mathrm{GP}}\)) measures success retention when distractors are added as a noise (i.e., the Level~\(\ell\) Gold-present condition). Higher Robustness indicates less disruption from distractors. We also report the mean and standard deviation of Robustness across Levels~1--5 to summarize overall stability.

\begin{table}[t]
\centering
\caption{
Level-wise Adaptability and Robustness metrics in percentages.
\textbf{Robustness} is reported level-wise across Levels~1--5, with the final column summarizing the mean $\pm$ standard deviation.
}
\label{tab:adaptability_robustness}
\small
\setlength{\tabcolsep}{4pt}
\begin{tabular}{lrrrrrrr}
\toprule
\textbf{Model} & \textbf{Adaptability} & \textbf{Rob.(L1)} & \textbf{Rob.(L2)} & \textbf{Rob.(L3)} & \textbf{Rob.(L4)} & \textbf{Rob.(L5)} & \textbf{Robustness Avg.} \\
\midrule
GPT-4o-mini & 59.68 & 89.95 & 89.64 & 89.44 & 90.84 & 89.61 & $89.90 \pm 0.56$ \\
Llama3-8B & 8.63 & 66.07 & 67.65 & 66.82 & 77.20 & 72.95 & $70.14 \pm 4.78$ \\
Qwen2.5-7B & 65.12 & 94.29 & 94.81 & 90.32 & 89.55 & 86.91 & $91.18 \pm 3.33$ \\
\bottomrule
\end{tabular}
\end{table}

Table~\ref{tab:adaptability_robustness} reveals clear model-specific differences.
Qwen2.5-7B achieves the highest Adaptability (65.12\%), suggesting it can effectively recover from missing gold tools through internal reasoning or alternative tools.
It also achieves the highest average Robustness (\(91.18 \pm 3.33\)), indicating strong resilience against distractors, despite a slight performance decline from Level~1 to Level~5. GPT-4o-mini also shows strong Adaptability (59.68\%).
While its absolute average Robustness is slightly lower than Qwen2.5-7B, it is extremely stable across distractor levels (\(89.90 \pm 0.56\)), showing consistent resilience to distractors even if its ability to recover without gold tools is slightly weaker. In contrast, Llama3-8B exhibits much weaker Adaptability (8.63\%) and substantially lower average Robustness (\(70.14 \pm 4.78\)).
Interestingly, its higher Robustness at Levels~4--5 suggests Llama3-8B is less penalized when distractors are highly similar to intended tools, though its overall retention remains poor.
Overall, these metrics demonstrate that \ToolMATH effectively evaluates how model success is preserved or disrupted under systematically altered tool availability and catalog difficulty.

\subsection{Tool Connectivity Across Models}
\label{sec:tool-connectivity}

Adaptability and Robustness evaluate how model success changes when the tool catalog changes.
We now isolate a complementary axis: whether models can preserve correctness as the executed tool-use trajectory becomes longer.
This captures \emph{tool connectivity}, i.e., the ability to chain tool calls, reuse intermediate observations, and maintain a coherent execution path across multiple dependent steps.

\begin{figure}[t]
    \centering
    \includegraphics[width=0.8\textwidth]{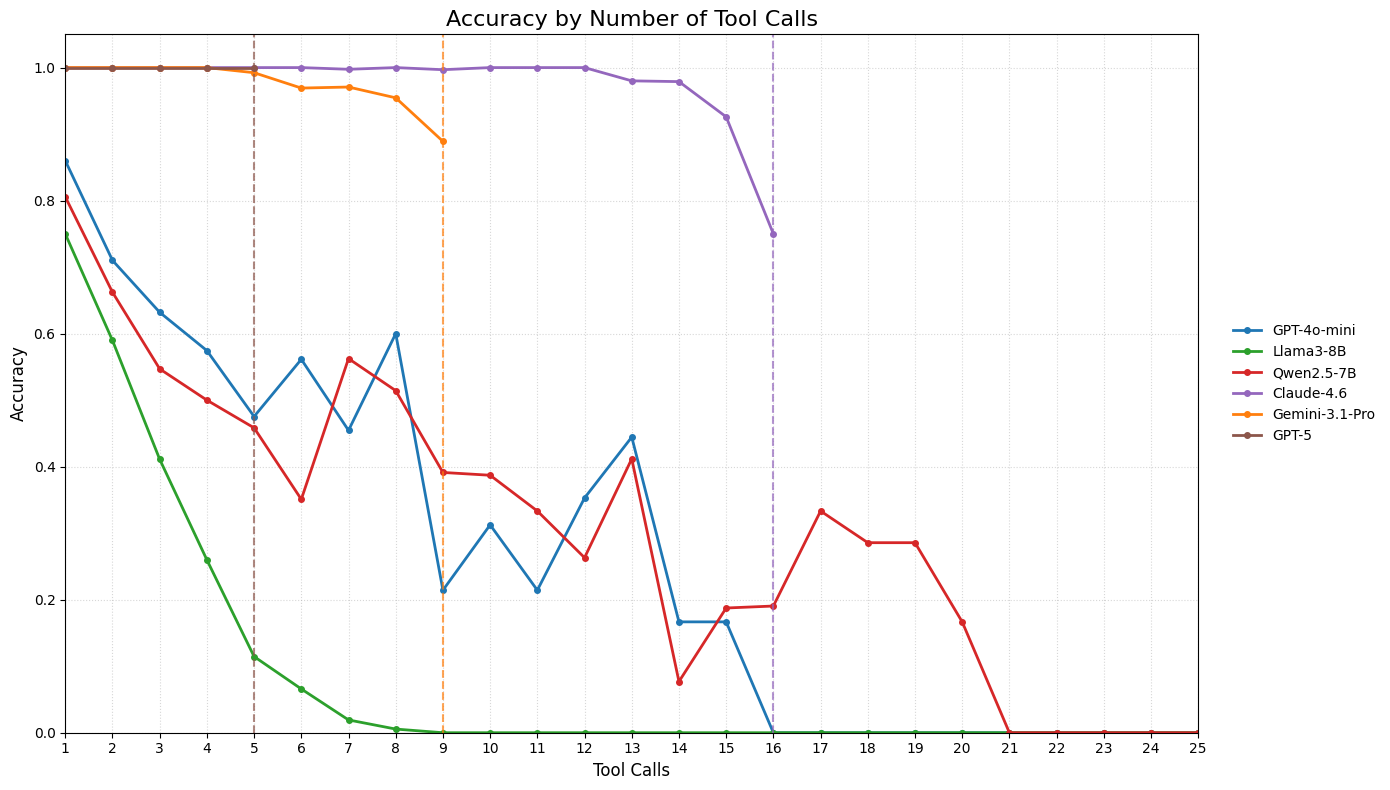}
    \caption{
    \textbf{Accuracy by the number of executed tool calls across models.}
    Accuracy is plotted as a function of tool-call count under the main Plan+ReAct protocol and the same distractor budget of \(k=5\) with Level~3 sampling.
    For shorter trajectories, vertical dashed lines mark the last observed tool-call count rather than implying failure beyond that point.
    }
    \label{fig:numtool-accuracy}
\end{figure}

Figure~\ref{fig:numtool-accuracy} shows that models differ not only in whether they call tools, but also in how well they remain connected across longer tool-use trajectories.
GPT-4o-mini, Llama3-8B, and Qwen2.5-7B degrade significantly as tool-call chains lengthen, indicating that later failures are not merely caused by choosing whether to use tools, but by maintaining a correct sequence of dependent calls.
Among these models, Llama3-8B shows the steepest decline, while GPT-4o-mini or Qwen2.5-7B preserves useful execution behavior over a wider range of trajectories before eventually breaking down.

By contrast, Claude-4.6, Gemini-3.1-Pro, and GPT-5 remain much more stable over their observed tool-call ranges.
Their curves suggest that stronger models are better able to preserve intermediate state, select subsequent tools consistently, and integrate tool observations into later reasoning steps.
The vertical dashed markers are important for interpretation: they indicate where each corresponding curve stops being observed, rather than where accuracy should be treated as zero.
GPT-5 is the most extreme case of this distinction.
Its observed curve stays at ceiling-level performance, but this does not necessarily mean that GPT-5 is the best model at executing long connected tool chains.
Rather, GPT-5 appears to adopt a highly conservative tool-use policy: when a problem would otherwise require a longer sequence of tool calls, it often solves the problem directly or minimizes tool use instead of extending the tool-call trajectory.
Thus, the absence of later observations should be read as part of the model's behavior---a tendency to avoid long tool-use chains---rather than as evidence that it would maintain the same accuracy if forced to execute them.

Overall, this analysis separates long-horizon execution reliability from the catalog-shift metrics above.
A model may be robust to distractors or adaptive when gold tools are removed, yet still fail once solving requires a longer connected chain of tool calls.
Tool Connectivity therefore provides a third diagnostic view of tool-use behavior, complementing Adaptability and Robustness before we turn to the harder setting where intended tool availability becomes more critical.

\subsection{\ToolMATHHard: tool availability and insufficiency}
\label{sec:main-hard}

Appendix~\ref{sec:results-splits} provides all-model curves comparing \ToolMATH and \ToolMATHHard; we summarize the main trend here.
\ToolMATHHard tests whether intended tool availability matters more on intrinsically difficult multi-step problems.
Unlike the main \ToolMATH split, \ToolMATHHard contains questions for which automatically extracted tools did not yield a validated tool-use trajectory; for these questions, we provide separately validated human-authored gold tools.

\ToolMATHHard shows a substantially larger gap between the Gold-only condition and settings without gold tools, namely the No tools baseline and the Distractors-only condition.
This gap is especially pronounced at higher logical hops, indicating that successful execution in the hard split depends more strongly on intended tool capabilities.
By contrast, the main \ToolMATH split often shows a smaller Gold-only vs.\ Distractors-only separation, suggesting that many instances can still be solved through internal reasoning or alternative tool-consistent paths.

\subsection{Trace-level failure diagnostics}
\label{sec:main-failure}

Appendix~\ref{sec:results-failurecases} details trace-level diagnostics; we summarize the main finding here.
Aggregate accuracy and Tool-Acc do not reveal how failures are organized inside a tool-use trace.
We therefore annotate failed traces with a non-exclusive taxonomy covering planning, tool selection, tool hallucination, parameterization, formatting, reasoning, observation use, repeated calls, and incomplete execution.

Co-occurrence analysis shows that tool-catalog conditions change not only failure frequencies, but also how failure modes couple.
In particular, tool selection errors become more broadly coupled with other errors when distractors are introduced or when gold tools are removed, suggesting that catalog reliability affects the internal organization of tool-use failures.
These results support the role of \ToolMATH as a diagnostic benchmark for failure structure, not merely final-answer accuracy.

\subsection{Framework sensitivity}
\label{sec:main-framework}

\ToolMATHHard also reveals sensitivity to the tool-use controller.
Under the Gold-only condition, we compare No tools, ReAct, DFSDT, and Plan+ReAct; detailed curves and heatmaps are provided in Appendix~\ref{sec:results-framework}.
The main trend is that controller differences are modest at low hops but become clearer at higher hops, where explicit planning more consistently preserves accuracy.
This indicates that \ToolMATH can diagnose not only model-level tool-use behavior, but also the effect of controller design on long-horizon tool execution.

\paragraph{Key takeaways.}
Overall, \ToolMATH exposes four complementary dimensions of tool-use behavior: long-horizon difficulty through logical-hop complexity, catalog sensitivity through distractor similarity and missing gold tools, adaptability/robustness under tool removal, and trace-level failure organization through co-occurrence diagnostics.
Together, these axes make \ToolMATH a diagnostic benchmark rather than a final-accuracy-only evaluation.

\section{Limitations}
\label{sec:limitations}

\ToolMATH is math-grounded, which enables deterministic scoring and controlled construction but does not cover the full ambiguity, underspecification, and open-ended objectives found in general tool-use domains.
Its tools are derived from stepwise MATH solutions, so the benchmark emphasizes symbolic tool use rather than interaction with real-world APIs, services, or dynamically changing environments.
Although our validation-and-repair pipeline reduces benchmark artifacts, it cannot fully eliminate extraction noise, source-dataset bias, or validation bias.
Finally, our behavioral metrics and trace-level co-occurrence analyses are diagnostic rather than causal: they reveal how model behavior changes under controlled tool-catalog shifts, but do not by themselves identify the underlying mechanism responsible for each failure.

\section{Conclusion}
\label{sec:conclusion}

We introduced \ToolMATH, a math-grounded diagnostic benchmark for evaluating long-horizon tool use under systematic tool-catalog constraints.
\ToolMATH constructs reusable hidden-code tools from stepwise MATH solutions and exposes models to controlled Gold-only, Gold-present, and Distractors-only conditions.
This design allows us to measure three complementary capabilities: Adaptability --- how much Gold-only success is preserved when the intended tools are removed, Robustness --- how much success is retained when distractors are added while gold tools remain available, and tool connectivity --- whether models can compose logically dependent tool calls across multiple tool calls.
Together with behavior-conditioned metrics and trace-level failure co-occurrence analysis, \ToolMATH shows that models with similar final accuracy can differ substantially in whether they avoid tools, use them reliably, over-call under missing capabilities, or fail through coupled planning and tool-selection errors.
By making tool availability, distractor similarity, and long-horizon dependency structure controllable, \ToolMATH provides a diagnostic testbed for studying not only whether tool-augmented models answer correctly, but how their success changes as the tool environment changes.





\clearpage
\appendix

\section{Gold-Present Difficulty Trends}
\label{app:gold-present-trends}

\begin{figure}[!h]
    \centering
    \includegraphics[width=0.68\columnwidth]{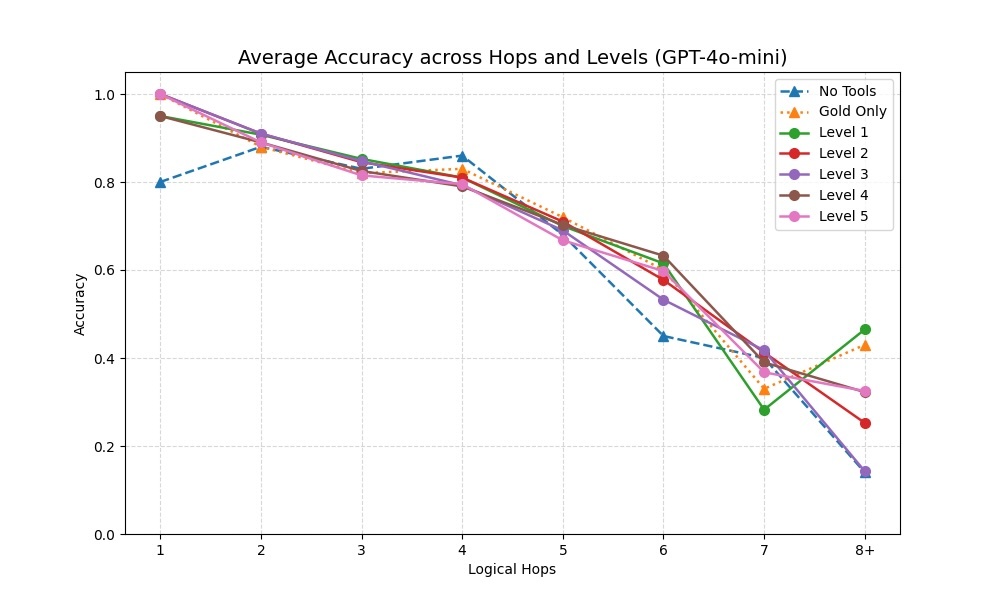}
    \caption{\textbf{GPT-4o-mini} accuracy by logical-hop group, averaged over distractor budgets \(k\).
    Curves show the No tools baseline, the Gold-only condition, and the Gold-present condition with distractor similarity Levels~1--5.
    Accuracy decreases with hop count; higher-similarity distractors separate most clearly in higher-hop regimes.}
    \label{fig:avg_gpt4o}
\end{figure}

We aggregate results from the Gold-present condition across distractor budgets \(k\) and plot accuracy as a function of logical-hop count, with separate curves for distractor similarity levels (Levels~1--5) and two baselines: the No tools baseline and the Gold-only condition.
Figure~\ref{fig:avg_gpt4o} shows GPT-4o-mini as a representative example; full heatmaps for the Gold-present condition across hop counts and distractor budgets are provided in Appendix~\ref{sec:appendix-gold}.

Accuracy generally decreases as hop count increases across all similarity levels, and the same trend holds for the No tools baseline.
This confirms that hop count captures intrinsic long-horizon, multi-step difficulty beyond effects from tool-set overlap alone.
Higher similarity levels are associated with larger performance differences at higher hops: while curves are close at low hops (hop~1--2), high-overlap levels show larger drops at hop~5+ and separate more clearly in the hardest regimes (hop~7 and hop~8+).
This interaction suggests that distractor similarity is most informative when evaluated together with long-horizon dependency structure, rather than as an isolated source of difficulty.

The Gold-only condition remains near the ceiling at low hops and stays above No tools at moderate hop depths, but performance still declines substantially at higher hops.
Thus, tool availability is a necessary but insufficient condition: long-horizon plan consistency and correct intermediate dependency tracking remain the bottleneck.
Overall, logical-hop complexity emerges as the dominant difficulty axis, with distractor similarity acting as an amplifying factor rather than the primary driver of failures.

\section{Full Gold-Present Heatmaps}
\label{sec:appendix-gold}

This appendix provides full heatmaps for the Gold-present condition (hop $\times$ distractor budget $k$, repeated across Levels~1--5) for the three evaluated models.

\begin{figure}[!h]
    \centering
    \includegraphics[width=0.75\textwidth]{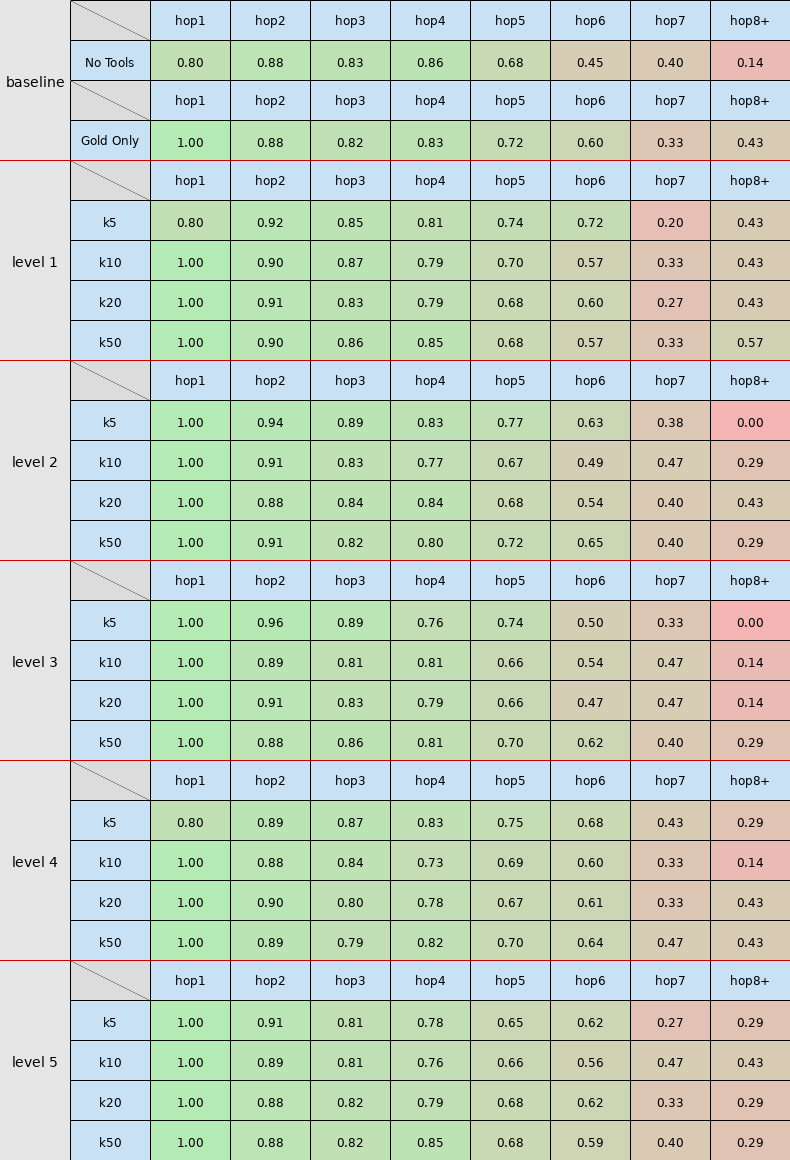}
    \caption{Gold-present condition results for \textbf{GPT-4o-mini} (full grid). The grid reports accuracy over logical hops (columns) and distractor set size $k$ (rows), for distractor sampling levels 1--5. Baselines (No tools and Gold-only) are shown at the top.}
    \label{fig:app_gold_gpt4o}
\end{figure}
\FloatBarrier

\begin{figure}[!h]
    \centering
    \includegraphics[width=0.75\textwidth]{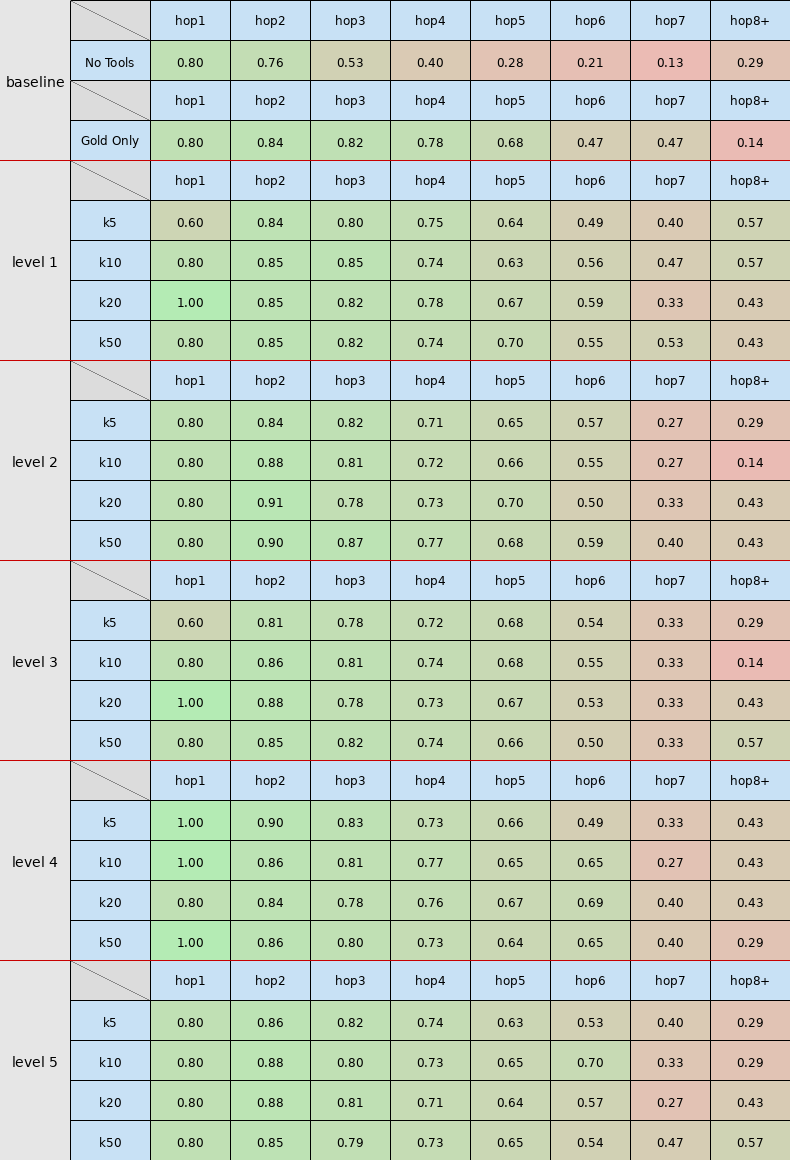}
    \caption{Gold-present condition results for \textbf{Qwen 2.5-7B} (full grid).}
    \label{fig:app_gold_qwen}
\end{figure}
\FloatBarrier

\begin{figure}[!h]
    \centering
    \includegraphics[width=0.75\textwidth]{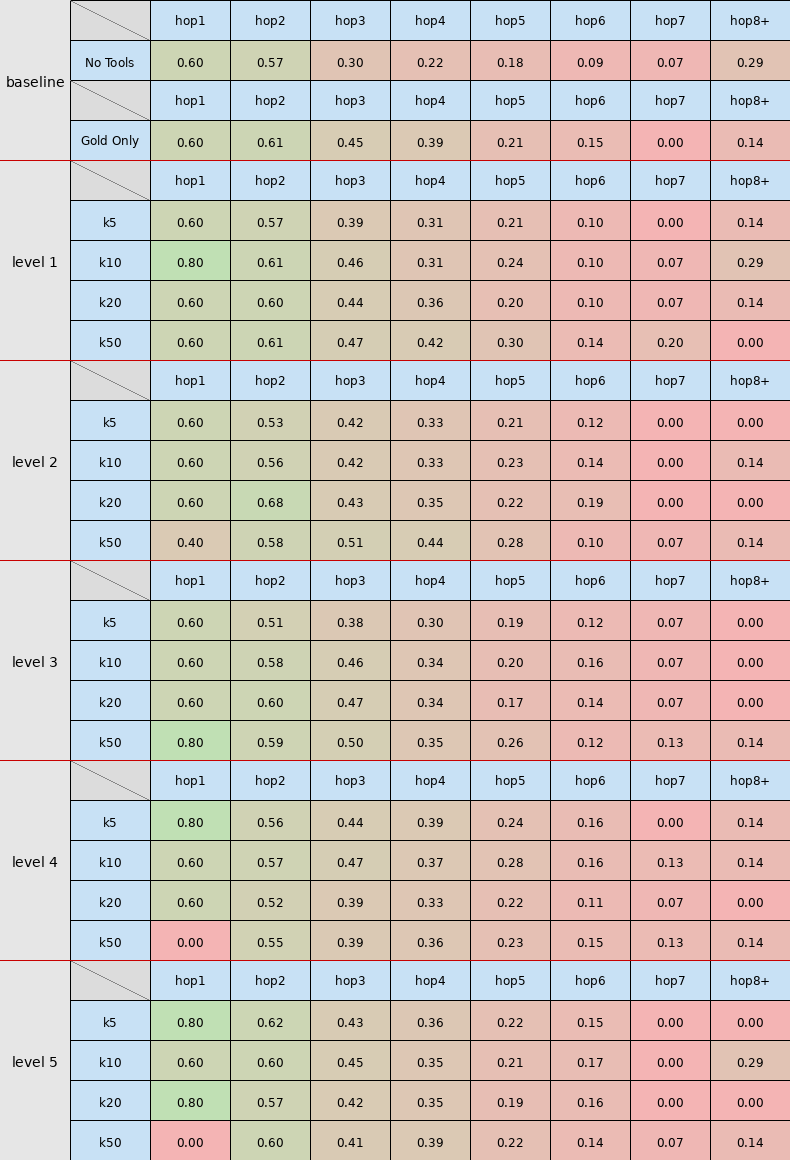}
    \caption{Gold-present condition results for \textbf{Llama 3-8B} (full grid).}
    \label{fig:app_gold_llama}
\end{figure}
\FloatBarrier

\section{Detailed evaluation results for the Distractors-only condition}
\label{sec:appendix-nogold}

\begin{figure}[!h]
    \centering
    \includegraphics[width=0.8\textwidth]{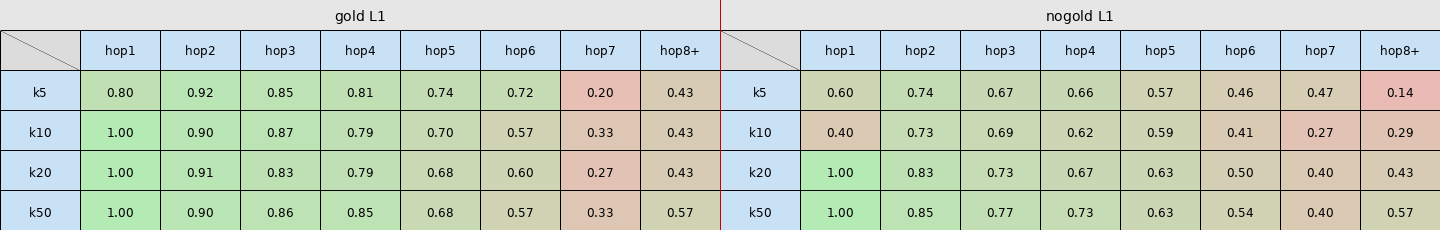}
    \caption{Gold-present condition (Level~1) vs.\ Distractors-only condition (Level~1) for \textbf{GPT-4o-mini} (full grid).}
    \label{fig:app_nogold_gpt4o}
\end{figure}
\FloatBarrier

\begin{figure}[!h]
    \centering
    \includegraphics[width=0.8\textwidth]{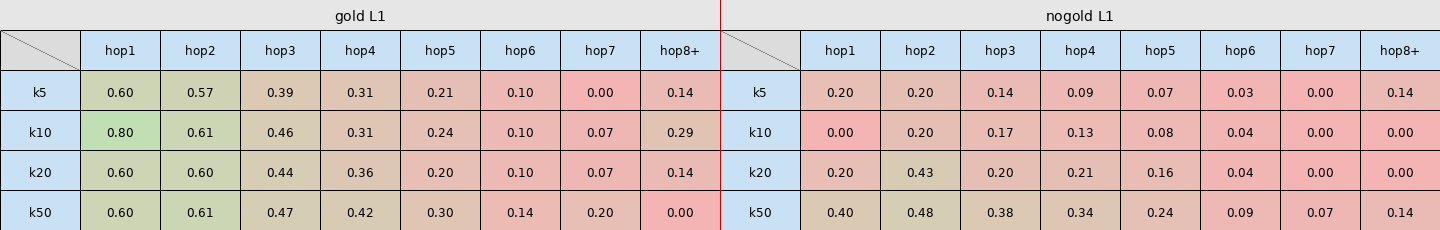}
    \caption{Gold-present condition (Level~1) vs.\ Distractors-only condition (Level~1) for \textbf{Llama 3-8B} (full grid).}
    \label{fig:app_nogold_llama}
\end{figure}
\FloatBarrier

\begin{figure}[!h]
    \centering
    \includegraphics[width=0.8\textwidth]{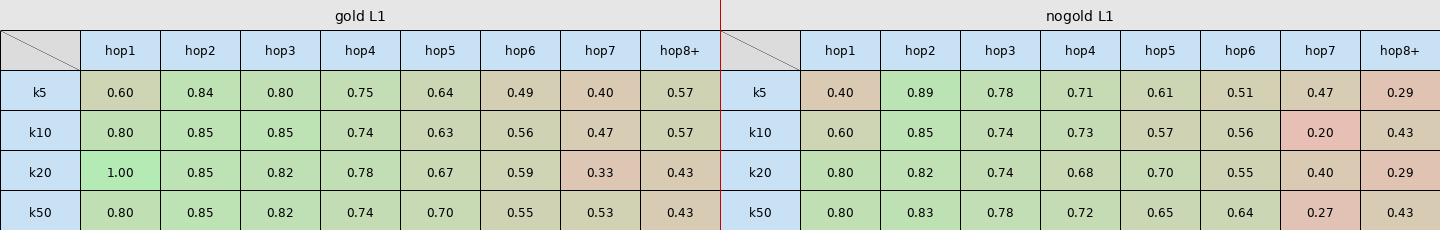}
    \caption{Gold-present condition (Level~1) vs.\ Distractors-only condition (Level~1) for \textbf{Qwen 2.5-7B} (full grid).}
    \label{fig:app_nogold_qwen}
\end{figure}
\FloatBarrier

\section{Detailed Framework Results for \ToolMATHHard}
\label{sec:results-framework}

This appendix reports detailed framework-sensitivity results on \ToolMATHHard under the Gold-only condition.
We compare four execution settings: No tools, ReAct, DFSDT, and Plan+ReAct.
The goal is to isolate whether performance differences arise only from the base model, or also from the controller used to organize multi-step tool execution.

\begin{figure}[!h]
    \centering
    \includegraphics[width=0.95\columnwidth]{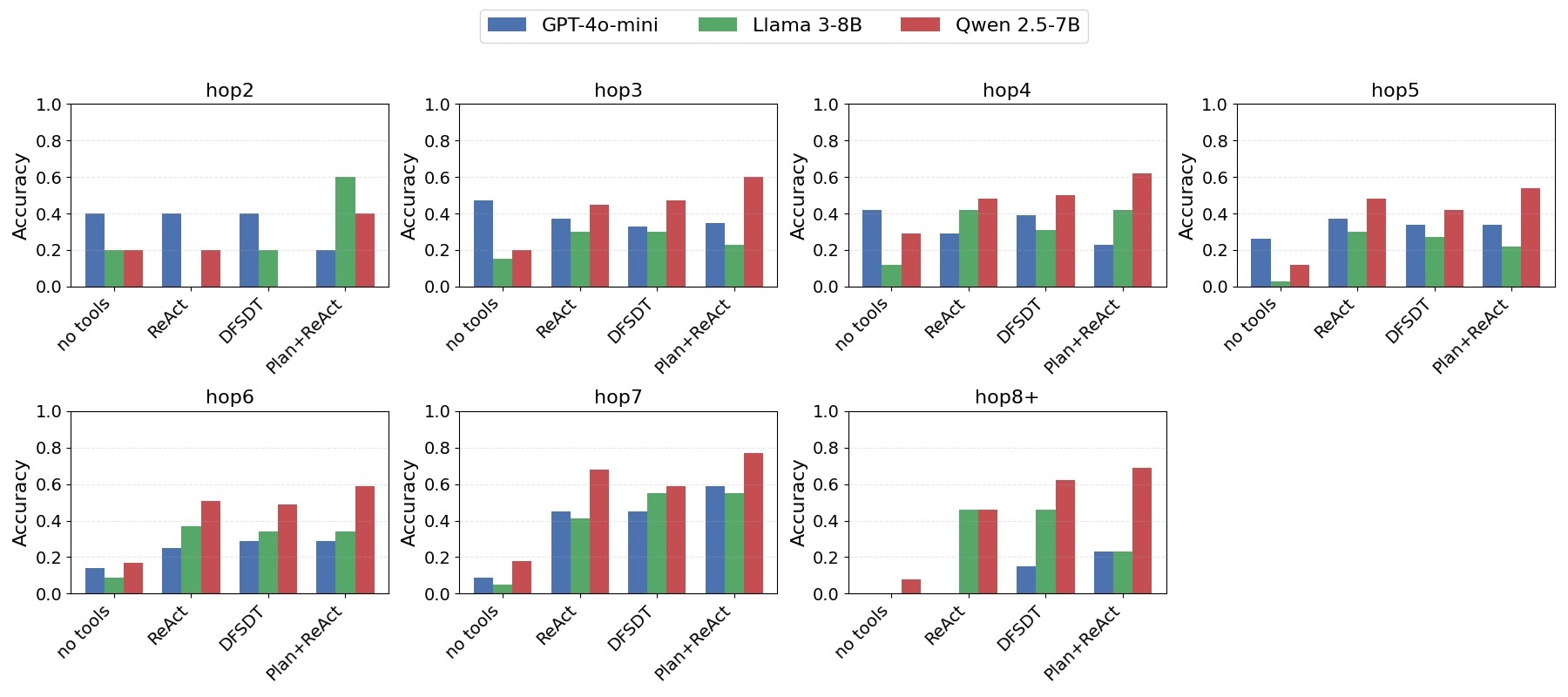}
    \caption{\textbf{Hop-wise framework comparison under Gold-only tool lists.}
    Framework differences are small at low hops but separate at higher hops, where Plan+ReAct most consistently maintains accuracy.}
    \label{fig:app-main-framework}
\end{figure}

Figure~\ref{fig:app-main-framework} summarizes the hop-wise framework comparison across models.
At lower hop counts, the differences among ReAct, DFSDT, and Plan+ReAct are often modest, suggesting that short tool-use traces can be handled by relatively simple interaction protocols.
At higher hop counts, the gap between controllers becomes clearer: explicit planning tends to maintain accuracy more consistently when later decisions depend on earlier tool observations.
DFSDT improves over ReAct in some intermediate-hop regimes, but is less consistently strong than Plan+ReAct at the highest hop depths.

\begin{figure}[!h]
    \centering
    \includegraphics[width=0.8\textwidth]{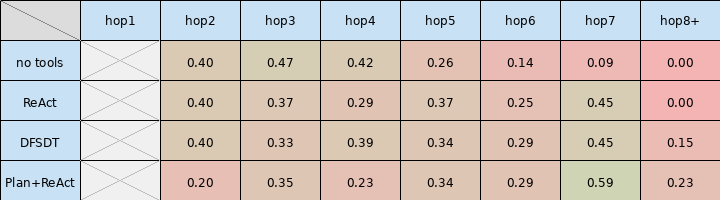}
    \caption{Hard-set frameworks for \textbf{GPT-4o-mini} (heatmap): no tools vs.\ ReAct vs.\ DFSDT vs.\ Plan+ReAct.}
    \label{fig:app_hardset_merged_gpt4o}
\end{figure}

\begin{figure}[!h]
    \centering
    \includegraphics[width=0.8\textwidth]{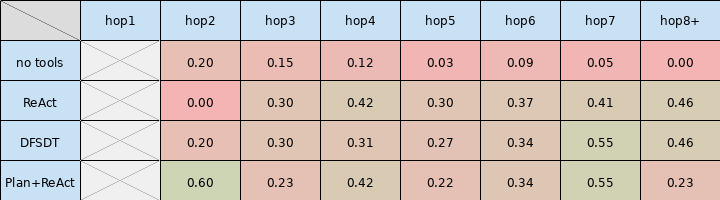}
    \caption{Hard-set frameworks for \textbf{Llama 3-8B} (heatmap): no tools vs.\ ReAct vs.\ DFSDT vs.\ Plan+ReAct.}
    \label{fig:app_hardset_merged_llama}
\end{figure}

\begin{figure}[!h]
    \centering
    \includegraphics[width=0.8\textwidth]{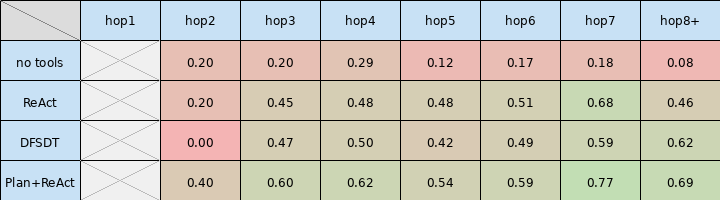}
    \caption{Hard-set frameworks for \textbf{Qwen 2.5-7B} (heatmap): no tools vs.\ ReAct vs.\ DFSDT vs.\ Plan+ReAct.}
    \label{fig:app_hardset_merged_qwen}
\end{figure}

Figures~\ref{fig:app_hardset_merged_gpt4o}--\ref{fig:app_hardset_merged_qwen} provide model-specific heatmaps.
They show that framework sensitivity is also model-dependent.
Llama 3-8B exhibits stronger degradation and larger dependence on the controller, while Qwen 2.5-7B benefits more consistently from Plan+ReAct in higher-hop regimes.
These results support using \ToolMATHHard as a diagnostic testbed for controller design in long-horizon tool use.

\FloatBarrier

\FloatBarrier

\FloatBarrier

\FloatBarrier

\FloatBarrier

\section{Tool Extraction and Logical-Hop Annotation Details}
\label{app:tool_extraction_and_hops}

This appendix describes (i) how we extract reusable tools from annotated MATH solutions, (ii) the tool schema and storage format, and (iii) how we annotate each problem with a logical-hop count used for evaluation.

\subsection{Tool Extraction Prompt}
\label{app:tool_extraction_prompt}

We extract tools by prompting a function generator model to return a list of reusable helper functions as a JSON array.
The prompt encourages multiple small functions (rather than monolithic solutions) and constrains parameter types to a fixed whitelist to simplify downstream tool schemas.

\noindent\textbf{Prompt template (verbatim).}
\begin{promptbox}
You are a Python function-generator for a math-solver system.

**Task**
Extract every *reusable* helper function that appears implicitly in the
following solved problem.  Each function should capture a general
mathematical operation.  Aim for **multiple small functions** if the
solution has several distinct steps.

**Output format**
Return a **JSON array**.  Each element is an object with **exactly** these keys:
- "name"        - concise snake_case identifier
- "description" - what the function does and parameter informations.
- "inputs"      - dict {"param": "<python_type>"}
                 allowed types: <WHITELISTED_TYPES>
- "function"    - full Python `def` (no explanations)

Example (array with two functions):
[
  {
    "name": "common_divisors",
    "description": "Return sorted list of common divisors of two integers a and b.",
    "inputs": {"a": "int", "b": "int"},
    "function": "def common_divisors(a: int, b: int):\\n    ...return divs"
  },
  {
    "name": "integer_points_in_interval",
    "description": "Number of integers in the open interval (lo, hi], where lo and hi are integers.",
    "inputs": {"lo": "int", "hi": "int"},
    "function": "def integer_points_in_interval(lo: int, hi: int):\\n    ...return count"
  }
]

If **no reusable function** can be identified, output an *empty array* `[]`.
Do NOT wrap the JSON in markdown fences.
---
Problem:
{problem}

Solution:
{solution}
\end{promptbox}

\subsection{Tool Schema and Storage Format}
\label{app:tool_schema}

Each extracted tool is represented by a tuple consisting of a tool identifier, natural-language documentation, a typed input schema, and a Python implementation.
For evaluation, models only observe the \emph{name}, \emph{description}, and \emph{input schema}; the Python code is hidden and executed by the environment.

In our stored JSON, each tool record includes provenance fields that link back to the originating MATH problem and its official solution.
In the stored artifact, the \texttt{function} field points to a separate \texttt{.py} file containing the implementation.

\noindent\textbf{Schema example (truncated).}
\begin{promptbox}
[
  {
    "name": "absolute_difference",
    "description": "Calculate the absolute difference between two integers. Parameters: x (int), y (int). Returns: int.",
    "inputs": {
      "x": "int",
      "y": "int"
    },
    "function": "ao5wa0.py",
    "source_problem": "Two arithmetic sequences $A$ and $B$ both begin with 30 ... What is the absolute value of the difference between the 51st term of $A$ and the 51st term of $B$?",
    "source_solution": "The $n$th term is $a_n=a_1+d(n-1)$. ... Substituting $n=51$ ... is $\\boxed{1000}$."
  },
  ...
]
\end{promptbox}

\subsection{Logical-Hop Annotation}
\label{app:hops}

We annotate each problem with a \emph{logical-hop} count intended to reflect the number of sequentially dependent reasoning/tool steps required.
Unlike a raw count of extracted tools, hop count attempts to discount steps that can be carried out in parallel.

\paragraph{Step extraction.}
For each (problem, official solution), we first prompt a model to extract a linearized list of high-level solution steps.
These steps are written in natural language and are not tied to any single tool implementation.

\paragraph{Parallelization check and hop construction.}
We then build hop groups iteratively.
Starting from the first step, we maintain a list of existing hop groups.
For each new step, we prompt a model to judge whether the step can be parallelized with any existing hop group (i.e., it does not require outputs from that group).
If it can be parallelized, we assign it to that group and do not increase the hop count.
Otherwise, we create a new hop group and increase the hop count by one.
In evaluation plots, we report hop groups hop~1--7 and bucket all hops $\ge 8$ as \texttt{hop8+}.

\noindent\textbf{Prompt for step extraction.}
\begin{promptbox}
You are a careful math-solution analyst.
Given the problem and the official solution, extract an ordered list of
high-level solution steps. Output JSON only:
{"steps": ["...", "...", ...]}
\end{promptbox}

\noindent\textbf{Prompt for parallelization check.}
\begin{promptbox}
You are analyzing whether a new solution step depends on prior steps.
Given:
(1) A list of existing hop groups (each group is a set of steps that can run in parallel)
(2) A candidate new step
Decide whether the candidate can be placed into any existing hop group
without depending on outputs from that group.
Output JSON only:
{"parallelizable": true/false, "group_index": <int or -1>}
\end{promptbox}

\section{Multi-model Tool-wise Validation Details}
\label{app:toolwise_system_prompt}

This appendix describes the multi-model tool-wise validation procedure used to check whether a tool's natural-language description and input schema match its executable behavior.
The validation task follows the original judgment-based protocol: validator models do not predict replacement outputs, but instead decide whether the implementation's actual output is correct for each test case.

\paragraph{Rule-based test input generation.}
For each tool, we generate five schema-valid inputs using rule-based templates according to the declared parameter types.
For numeric parameters, we include small integers, boundary-like values, and randomly sampled values within safe execution ranges.
For list-valued parameters, we sample short lists with valid element types and include edge cases such as singleton or repeated-value lists when applicable.
The tool implementation is then executed on each input to obtain the actual output.

\paragraph{Judgment-based validation.}
Each validator model receives the tool description, input schema, concrete inputs, and actual outputs returned by the implementation.
For each test case, the validator judges whether the actual output is correct according to the described semantics of the tool.
We use three validator models: GPT-4o, Llama 3-8B, and Qwen 2.5-7B.

\paragraph{Majority validation rule.}
For each test input \(x_j\), let \(y_j=f_t(x_j)\) denote the output produced by the tool implementation.
Each validator model \(m\) produces a binary judgment
\[
z_{j}^{(m)} \in \{\mathrm{correct},\mathrm{incorrect}\},
\]
where the judgment indicates whether \(y_j\) is consistent with the tool description and input schema.
We aggregate the three judgments by majority vote:
\[
z_j=\mathrm{Majority}\left(
z_{j}^{(\mathrm{GPT\text{-}4o})},
z_{j}^{(\mathrm{Llama3})},
z_{j}^{(\mathrm{Qwen2.5})}
\right).
\]
A test case passes if
\[
z_j=\mathrm{correct}.
\]
A tool is retained only if all five test cases pass.
We allow small floating-point tolerance and benign formatting differences, such as \texttt{int} versus \texttt{float} when numerically equal, or list versus tuple for the same sequence.

\paragraph{Validator prompt.}
\begin{promptbox}
You are validating the behavior of a small deterministic math utility function.
You are given its natural-language description and several testcases.
Each testcase has:
- concrete inputs (already converted to the right Python types), and
- the actual_output returned by the implementation.

For EACH case, you must decide whether the actual_output is correct,
based on what the function is *supposed* to do from the description.

Guidelines:
- Carefully infer the intended behavior from the description (and input schema if needed).
- Compute or reason about the correct output for each testcase.
- Compare that internal expected result to actual_output.
- Allow small floating-point rounding errors (~1e-6) and harmless formatting differences
  (e.g., extra parentheses, int vs float when numerically equal, tuple vs list for the same sequence).
- If the implementation's result is clearly wrong (wrong sign, wrong magnitude, wrong structure),
  mark the case as incorrect.
\end{promptbox}

\subsection{Validation and Final Dataset Composition}
\label{app:final-dataset}

\ToolMATH applies a sequential validation-and-repair pipeline to ensure that benchmark errors are not conflated with model errors.

\paragraph{Step 1: Tool-wise validation.}
We validate each extracted tool in isolation to ensure that its natural-language description and input schema are consistent with its executable behavior.
For every tool, we generate five schema-valid test inputs using a rule-based input generator that covers nominal, boundary, and randomly sampled values for the declared input types, and execute the tool implementation to obtain actual outputs.
We then provide the tool metadata (description and input schema), the concrete inputs, and the actual outputs to three validator models, \textbf{GPT-4o}, \textbf{Llama 3-8B}, and \textbf{Qwen 2.5-7B}.
Each validator judges whether the actual output is correct according to the described tool semantics.
For each test case, we aggregate the three binary judgments by majority vote.
A test case passes if the majority judgment is correct, and a tool passes tool-wise validation only if all five test cases pass.
Tools whose outputs are judged incorrect by the majority on at least one test case are discarded from the tool-wise validated set and handled by the downstream repair/audit stage.
This procedure reduces reliance on a single LLM judge while preserving the original validation task: validators do not generate replacement answers, but instead judge whether the implementation output matches the natural-language specification.
The validation prompt and normalization details are provided in Appendix~\ref{app:toolwise_system_prompt}.

\paragraph{Step 2: Question-wise validation (trace-based).}
After tool-wise validation, we perform question-wise validation to identify tools that are \emph{empirically usable} under our tool-use protocol, and to filter questions that admit at least one successful tool invocation in a correct solution.
For each MATH problem $p$, we provide each validator model with (i) the problem statement and (ii) its gold tool set $\mathcal{G}(p)$ (tool name, description, and input schema; code hidden), and run the same Plan+ReAct prompt that will be further used in the main evaluation.
A tool $t \in \mathcal{G}(p)$ is marked as \emph{question-wise validated} if there exists at least one validator model that (a) produces a correct final answer for $p$ and (b) successfully calls $t$ at least once in its trace (i.e., selecting $t$ from the provided list, supplying schema-valid arguments, and executing without runtime failure).
A question $p$ is marked as \emph{validated} if at least one tool in $\mathcal{G}(p)$ becomes validated by this criterion.
This trace-based criterion reflects our intended setting where models may combine internal reasoning with tool execution, rather than relying on tools for every step.

\paragraph{Validator set.}
We use a fixed validator set: $\mathcal{M}=$ \{GPT-4o-mini, Llama 3-8B, Mistral-7B, Qwen2-7B, Qwen 2.5-7B, Phi-3 Medium, and Yi 1.5-9B\}.

Concretely, let $\mathcal{C}(p,m)$ be the set of tools successfully executed by model $m$ while solving $p$ correctly.
Then the question-wise validated tool set is
\[
\mathcal{T}_{\mathrm{qw}} \;=\; \bigcup_{p}\;\bigcup_{\substack{m \in \mathcal{M}\\ \mathrm{Correct}(p,m)=1}}\;\mathcal{C}(p,m),
\]
and a question is retained if $\mathcal{G}(p)\cap \mathcal{T}_{\mathrm{qw}} \neq \emptyset$.

\paragraph{Step 3: Targeted human audit and tool repair.}
For episodes rejected by question-wise validation, we perform targeted human inspection.
If an episode is solvable, we reinstate it.
Otherwise, we revise the responsible tool(s) by editing descriptions and/or implementations, then re-run tool-wise validation for the modified tools and re-run question-wise validation for affected episodes.
Episodes that remain unsolved after this repair loop are excluded from the final release.

\section{Distractor Sampling Procedure and Safeguards}
\label{app:distractor_sampling}

This appendix details how we construct distractor tool lists for each problem and difficulty level (Levels 1--5), how we enforce nested inclusion across distractor budgets \(k\), and what safeguards prevent accidental gold overlap and tool-identity collisions.

\subsection{Global pool and per-problem exclusions}
\label{app:distractor_pool}

\paragraph{Global pool.}
All distractors are sampled from the same global pool \(\mathcal{U}\), defined as the union of \emph{filtered and validated} tools (i.e., tools that survive the tool-wise and question-wise filtering/repair pipeline). This ensures distractors are executable and description-consistent under our validation criteria.

\paragraph{Gold removal / self-exclusion.}
For each problem \(p\), let \(\mathcal{G}(p)\subset \mathcal{U}\) be its gold tool set.
We form an exclusion pool
\[
\mathcal{U}_{\setminus p} \;=\; \mathcal{U}\setminus \mathcal{G}(p),
\]
implemented by removing any tool instance whose provenance matches \(p\)'s gold tool instances (using a deterministic tool identifier derived from the tool's metadata; see \cref{app:distractor_safeguards}).
All distractor levels sample only from \(\mathcal{U}_{\setminus p}\), preventing accidental inclusion of gold tools.

\paragraph{Problem category.}
We assign each problem a broad category \(c(p)\) (e.g., Algebra, Geometry) using the dataset-provided \texttt{type} field.
Each tool inherits a category from its \texttt{source\_problem}.
We use \(c(\cdot)\) only to define candidate pools for Levels 1 and 3.

\subsection{Level 1--3: category-controlled random sampling}
\label{app:distractor_l123}

For each problem \(p\), we precompute an \emph{ordered} distractor list of length 100 for each level.
Randomness is reproducible via a fixed global seed and a fixed serialized order of \(\mathcal{U}\).

\paragraph{Candidate pools.}
We define:
\[
\mathcal{U}^{\text{same}}_{\setminus p} \;=\; \{ t\in \mathcal{U}_{\setminus p} : c(t)=c(p)\},\qquad
\mathcal{U}^{\text{diff}}_{\setminus p} \;=\; \{ t\in \mathcal{U}_{\setminus p} : c(t)\neq c(p)\}.
\]

\paragraph{Level 1 (exclude same-category).}
We sample 100 tools uniformly from \(\mathcal{U}^{\text{diff}}_{\setminus p}\).
If \(\mathcal{U}^{\text{diff}}_{\setminus p}\) is too small, we fall back to sampling from \(\mathcal{U}_{\setminus p}\) to ensure a full-length list.

\paragraph{Level 2 (pure random).}
We sample 100 tools uniformly from \(\mathcal{U}_{\setminus p}\).

\paragraph{Level 3 (same-category).}
We sample 100 tools uniformly from \(\mathcal{U}^{\text{same}}_{\setminus p}\).
If \(\mathcal{U}^{\text{same}}_{\setminus p}\) is too small, we fall back to \(\mathcal{U}_{\setminus p}\).

\paragraph{With/without replacement.}
If the chosen pool has at least 100 candidates, we sample \emph{without} replacement.
Otherwise, we sample \emph{with} replacement (cycling as needed) to produce exactly 100 distractors per level.

\subsection{Level 4: embedding similarity retrieval (top-100)}
\label{app:distractor_l4}

Level 4 selects distractors that are semantically close to the gold tools.

\paragraph{Embedding representation.}
Each tool \(t\) is embedded using the OpenAI embedding model \texttt{text-embedding-3-large}.
The embedded text is the concatenation of tool name and description:
\[
x(t) \;=\; \texttt{name}(t)\;\Vert\;\texttt{description}(t).
\]
Embeddings are \(\ell_2\)-normalized, and we use a cosine similarity.

\paragraph{Gold-to-candidate similarity score.}
For a candidate \(t \in \mathcal{U}_{\setminus p}\), we define its similarity to the gold set as
\[
s_4(t; p) \;=\; \max_{g \in \mathcal{G}(p)} \cos\!\big(e(t), e(g)\big),
\]
where \(e(\cdot)\) is the normalized embedding.
We then rank all candidates in \(\mathcal{U}_{\setminus p}\) by \(s_4(t;p)\) in descending order and take the top 100 as the ordered Level-4 distractor list.
If \(|\mathcal{U}_{\setminus p}|<100\), we cycle the ranked list to reach length 100.

\paragraph{Caching and efficiency.}
We cache tool embeddings and optionally cache top-\(K\) nearest neighbors per tool to avoid recomputing full similarity against the entire pool.
This does not change the ranking definition above; it only accelerates computation.

\subsection{Level 5: keyword overlap + embedding similarity (lexical-first ranking)}
\label{app:distractor_l5}

Level 5 prioritizes tools that are both lexically and semantically similar to the gold set.

\paragraph{Keyword extraction.}
We define a fixed keyword vocabulary of common mathematical terms (optionally extended with a small manual list).
For each tool \(t\), we extract keywords \(K(t)\) by tokenizing \(\texttt{name}(t)\) and \texttt{description}(t) and retaining tokens that appear in the vocabulary.
We also form the gold keyword union \(K(\mathcal{G}(p))=\bigcup_{g\in \mathcal{G}(p)} K(g)\).

\paragraph{Lexical overlap score.}
For each candidate \(t\in \mathcal{U}_{\setminus p}\), define
\[
o(t;p) \;=\; \big|K(t)\cap K(\mathcal{G}(p))\big|.
\]

\paragraph{Two-stage ranking.}
We rank candidates by the tuple \((o(t;p),\, s_4(t;p))\) in descending order (lexical overlap first, embedding similarity as tie-breaker).
Candidates with zero overlap remain eligible but are ranked below candidates with positive overlap.
We then take the top 100 (cycling if needed) as the ordered Level-5 distractor list.

\subsection{Nested distractor guarantee across budgets \(k\)}
\label{app:distractor_nested}

For each problem \(p\) and level \(\ell\in\{1,2,3,4,5\}\), we precompute an ordered list of 100 distractors,
\[
\mathbf{L}_{\ell}(p) = \big[t_1, t_2, \dots, t_{100}\big].
\]
For a distractor budget \(k \in \{5,10,20,50\}\), we define
\[
\mathcal{D}_{\ell,k}(p) \;=\; \{t_1,\dots,t_k\}.
\]
This construction guarantees \emph{nested inclusion}:
if \(k_1 < k_2\), then \(\mathcal{D}_{\ell,k_1}(p)\subseteq \mathcal{D}_{\ell,k_2}(p)\).
For Levels 1--3, the list order is fixed by a deterministic RNG seed; for Levels 4--5, the order is fixed by deterministic ranking.

\subsection{Safeguards and collision handling}
\label{app:distractor_safeguards}

\paragraph{No accidental gold overlap.}
We explicitly remove \(\mathcal{G}(p)\) from the candidate pool before any sampling or retrieval.
Implementation-wise, each tool instance is tracked by a deterministic identifier derived from tool metadata (including name, description, and implementation reference), and gold-instance identifiers are excluded from the distractor pool for the same problem.

\paragraph{Name collision safeguard (suffixing).}
Some extracted tools can share the same base name.
To avoid identity collisions in tool registries and to ensure that retrieval/sampling treats tools as distinct callable actions, we enforce unique tool names by appending a suffix, e.g.,
\texttt{solve\_linear\_a}, \texttt{solve\_linear\_b}, \dots\ .
This safeguard is applied before constructing distractor pools and remains consistent throughout evaluation.

\paragraph{Determinism and reproducibility.}
All randomized sampling uses a fixed global seed and a fixed serialized ordering of the validated tool pool.
Together with cached embeddings and deterministic sorting, this makes distractor lists reproducible across runs given the same validated tool pool.

\section{Tool-Use Protocol Prompts and Execution Rules}
\label{app:protocol_prompts}

This appendix specifies the exact prompts and execution rules used to compare \textbf{ReAct}, \textbf{Plan+ReAct}, and \textbf{DFSDT} on \ToolMATH, including (i) system/user prompts, (ii) tool-call formatting constraints, (iii) stopping criteria and budgets, (iv) retry/diversity rules, and (v) decoding and guardrails.

\subsection{Common tool-call interface and formatting}
\label{app:protocol_common}

\paragraph{Tool visibility.}
In all protocols, the model is provided with a list of available tools, each containing \emph{name}, \emph{description}, and a typed \emph{input schema}. Tool implementations are hidden and executed by the environment.

\paragraph{Action JSON constraints.}
When a tool is called, the model must emit exactly one \texttt{Action:} line containing a JSON object of the form:
\[
\texttt{\{ "name": "<tool\_name>", "arguments": \{...\} \}}
\]
The \texttt{arguments} object must be valid JSON and must conform to the tool's input schema (required keys present; value types match the declared schema). The environment rejects malformed JSON or schema-invalid arguments.

\paragraph{Observation.}
After an \texttt{Action} line, the model must wait for the environment to return an \texttt{Observation:} line containing the tool output. The model must condition subsequent reasoning and actions on this observation.

\paragraph{Duplicate-call guardrail.}
Across protocols, we disallow repeating the exact same \((\texttt{name}, \texttt{arguments})\) pair. If the model has already executed an identical call earlier in the same episode, it must reuse the prior observation rather than calling the tool again. This is enforced both as an instruction and by a trace-level cache used in evaluation.

\paragraph{Answer format.}
All protocols terminate by outputting a single line:
\[
\texttt{ANSWER: <numeric answer>}
\]
We use exact-match answer normalization for scoring (Section~\ref{sec:conditions}).

\subsection{ReAct-only protocol}
\label{app:protocol_react}

ReAct-only runs without an explicit planning stage. The model interleaves one-line thoughts with optional tool calls.

\paragraph{System prompt (verbatim).}
\begin{promptbox}
You are an expert competition-math solver.
For every turn output *exactly one* line beginning with "Thought: ...".
If you need a tool, follow IMMEDIATELY with ONE line:
Action: { "name": "<tool>", "arguments": { ... } }
Then WAIT for the Observation line before thinking again.
Never call a tool with the same arguments twice - reuse the prior Observation.
If you have already called a tool with the same name and identical arguments, DO NOT call it again; reuse the cached Observation verbatim and move forward.
Finish with:
ANSWER: <numeric answer>
\end{promptbox}

\subsection{Plan+ReAct protocol}
\label{app:protocol_planreact}

Plan+ReAct consists of two stages: (i) a planner that produces a short structured plan, and (ii) a ReAct-style solver that follows the plan while interacting with tools.

\paragraph{Planner system prompt (verbatim).}
\begin{promptbox}
You are an expert planner for competition-math problems.
Given the problem and available tools, write a concise, structured step-by-step plan.
Use a numbered list (1-7 steps). If a tool is needed, mention its exact name.
Do not solve the problem or compute final values.
Do not include any numeric answers or intermediate calculations.
Do not include explicit numbers, computed values, or worked examples; refer abstractly to the needed computations.
\end{promptbox}

\paragraph{Solver system prompt (verbatim).}
\begin{promptbox}
You are an expert competition-math solver.
For every turn output *exactly one* line beginning with "Thought: ...".
If your next action or tool choice deviates from the provided plan, include the token <PLAN_CHANGED>
in the Thought line for that turn (only when a deviation happens).
If you need a tool, follow IMMEDIATELY with ONE line:
Action: { "name": "<tool>", "arguments": { ... } }
Then WAIT for the Observation line before thinking again.
Never call a tool with the same arguments twice - reuse the prior Observation.
If you have already called a tool with the same name and identical arguments, DO NOT call it again; reuse the cached Observation verbatim and move forward.
Finish with:
ANSWER: <numeric answer>
\end{promptbox}

\subsection{DFSDT protocol (decision-tree search)}
\label{app:protocol_dfsdt}

DFSDT follows the decision-tree search style popularized by ToolLLM-like frameworks, where the model proposes a single action at a time and the controller branches over alternatives. Our prompts are adapted from ToolLLM (Appendix A.8) with minimal changes for math tool use.

\paragraph{System prompt (solver; verbatim).}
\begin{promptbox}
You are MathTool-GPT, capable of using provided functions to solve math problems.

Workflow:
1) I will give you a math problem.
2) At each step, analyze the current state and choose the next action by calling ONE function.
3) You will receive the function result (observation), then repeat.
4) When you have enough to answer, call Finish with return_type="give_answer" and include a complete final_answer.
5) If you cannot proceed effectively, call Finish with return_type="give_up_and_restart".

Rules:
- Keep your thoughts concise (max 5 sentences).
- Prefer tool calls for computation, algebra, simplification, verification.
- Make one attempt per idea; don't repeat the same call with the same arguments.
\end{promptbox}

\paragraph{User prompt (problem wrapper; verbatim).}
\begin{promptbox}
Task description:
Solve the following problem using the available functions.

Problem:
{problem_text}
\end{promptbox}

\paragraph{Diversity user prompt (different child node; verbatim).}
\begin{promptbox}
This is not the first time you try this state; previous attempts from here failed.

Here are previous action candidates you already tried from THIS STATE:
{previous_candidate_actions}

Now you MUST choose a next action that is different from all previous candidates above.
First analyze why they failed, then take a new action.
\end{promptbox}

\subsection{Stopping criteria, budgets, and retry rules}
\label{app:protocol_budgets}

\paragraph{Termination.}
For ReAct and Plan+ReAct, an episode terminates when the model outputs a line of the form
\texttt{ANSWER: <numeric answer>}.
For DFSDT, the controller terminates when the model calls \texttt{Finish} with \texttt{return\_type="give\_answer"} and provides a complete \texttt{final\_answer}.
If a run ends without producing a valid final answer format, it is counted as \textbf{no answer} in trace analysis.

\paragraph{Step budget.}
For ReAct and Plan+ReAct, we cap the solver loop at \textbf{16} model steps (i.e., up to 16 assistant generations in the tool-use loop after initialization).
Since each step permits at most one \texttt{Action} line, this also bounds the number of \emph{executed} tool calls by 16 (duplicate calls are intercepted by a cache and do not re-execute the tool).

\paragraph{Wall-clock timeouts.}
We enforce a per-question wall-clock timeout of \textbf{120 seconds}.
Separately, each tool execution is protected by a per-call timeout of \textbf{60 seconds}; if exceeded, the observation is set to an error message and the run continues until termination or the question timeout is reached.

\paragraph{Duplicate-call cache and enforcement.}
We cache tool results keyed by \((\texttt{name}, \texttt{arguments})\), using a canonical serialization of \texttt{arguments} (sorted keys and normalized numeric forms).
If the model repeats an identical call, the environment returns a cached observation instead of re-executing the tool and issues an explicit reminder not to repeat identical actions.
After multiple repeated identical actions, further identical actions are ignored and the model is instructed to proceed to the next reasoning step or output \texttt{ANSWER}.

\paragraph{Action parsing robustness.}
If the emitted \texttt{Action} JSON is malformed, the environment returns an explicit error observation and continues the loop.
If the model produces repeated non-progressing steps (e.g., repeated malformed actions or repeated duplicates), the environment injects a short reminder to follow the one-\texttt{Thought}-per-turn constraint and to reuse cached observations.

\paragraph{LLM call retries (API robustness).}
All LLM calls (planner, solver, and judge) use a retry wrapper that attempts up to \textbf{5} retries on transient API errors (e.g., rate limits or timeouts) with exponential backoff (base \textbf{0.8}) and jitter.
If retries are exhausted, the episode is terminated and counted as incorrect/no-answer depending on whether a valid final answer was produced.

\paragraph{DFSDT restarts and diversity.}
In DFSDT, when the model emits \texttt{Finish(return\_type="give\_up\_and\_restart")}, the controller restarts from the root state under the same global budgets.
When revisiting the same search state, the controller provides the diversity prompt listing previously tried action candidates from that state and requires the next proposed action to differ from them.

\subsection{Decoding, temperatures, and guardrails}
\label{app:protocol_decoding}

\paragraph{Decoding / temperature.}
We use near-deterministic decoding for the solver across protocols with \textbf{temperature \(=0.0\)} to reduce variance in execution traces.
In Plan+ReAct, the planner uses \textbf{temperature \(=0.2\)} to encourage concise but non-degenerated plan drafts, while the subsequent solver remains at \textbf{temperature \(=0.0\)}.

\paragraph{Formatting guardrails.}
All protocols require strict adherence to the one-line \texttt{Thought: ...} format per step.
If a tool is needed, the model must emit exactly one additional line
\texttt{Action: \{ "name": ..., "arguments": \{...\} \}}
and then wait for the environment \texttt{Observation}.
The environment validates that the \texttt{Action} payload is parseable (JSON-first, with a permissive fallback) and returns explicit error observations when parsing fails.

\paragraph{Execution guardrails.}
We enforce (i) per-tool and per-question timeouts, and (ii) cache-based prevention of exact duplicate tool executions, as described in Appendix~\ref{app:protocol_budgets}.

\section{Additional Results: \ToolMATH\ vs.\ \ToolMATHHard\ (All Models)}
\label{sec:results-splits}

This appendix provides the full all-model comparison between \ToolMATH and \ToolMATHHard under tool availability and insufficiency.
We compare the No tools baseline, the Gold-only condition, and the Distractors-only condition across logical-hop groups.
For all settings involving distractors, we use pure random sampling (Level~2) with \(k=10\) distractors.

\begin{figure}[!h]
    \centering
    \includegraphics[width=0.8\textwidth]{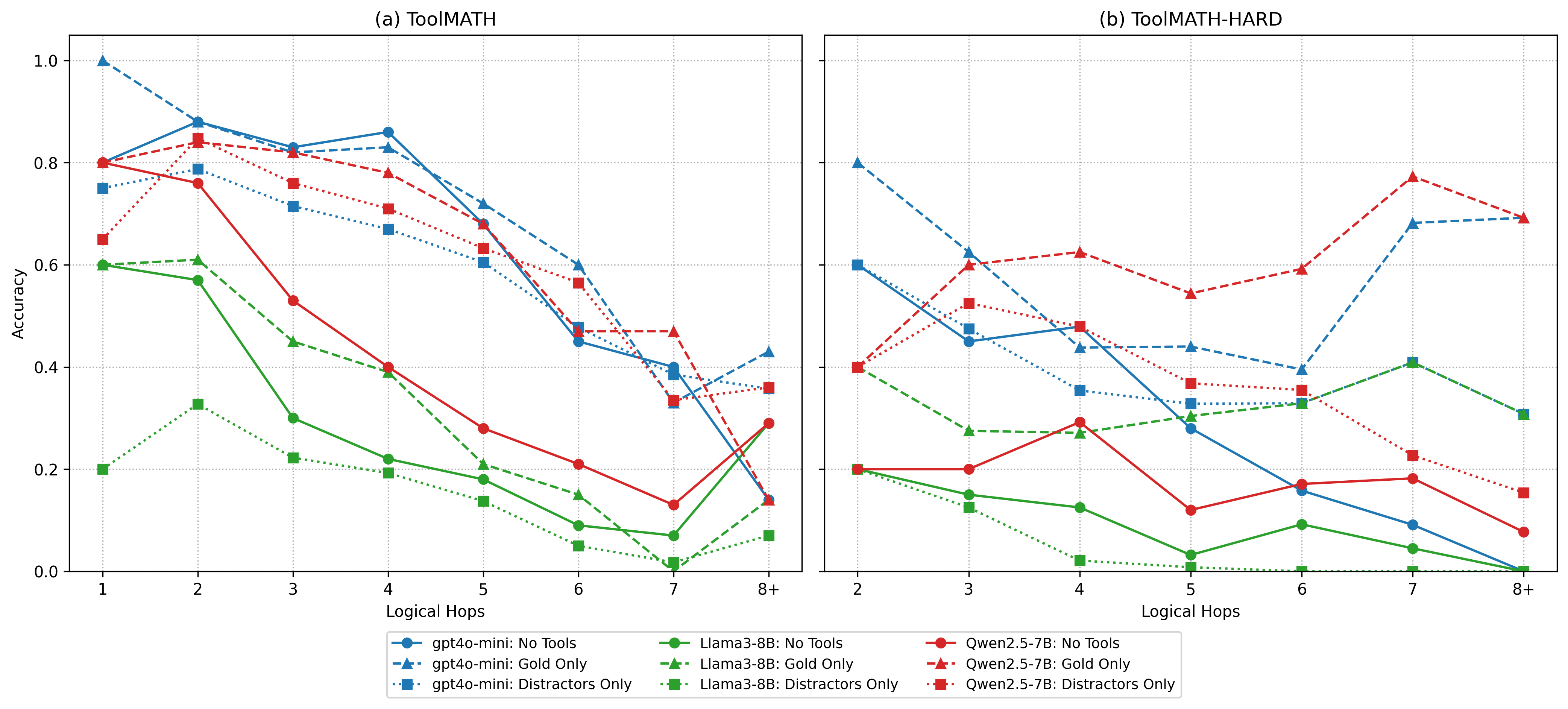}
    \caption{\textbf{\ToolMATH\ vs.\ \ToolMATHHard: hop-wise accuracy under tool availability and insufficiency (all models).}
    Left: \ToolMATH. Right: \ToolMATHHard.
    We report accuracy by logical-hop group under the No tools baseline, the Gold-only condition, and the Distractors-only condition.
    For all settings involving distractors, we fix the distractor list to pure random sampling (Level~2) with \(k=10\) tools.}
    \label{fig:app_normal_hard_merged_all}
\end{figure}

Figure~\ref{fig:app_normal_hard_merged_all} shows that the role of intended gold tools differs substantially between the main and hard splits.
On the main \ToolMATH split, the Gold-only condition usually improves performance over settings without gold tools, but the separation from the Distractors-only condition is often modest.
This suggests that many main-split examples remain solvable through internal reasoning or alternative tool-consistent trajectories.

On \ToolMATHHard, the Gold-only condition separates much more clearly from both the No tools baseline and the Distractors-only condition, especially at higher logical hops.
This indicates that the hard split contains cases where successful long-horizon execution relies more directly on the intended tool capabilities.
The larger gap also supports the intended role of \ToolMATHHard as a stress test for tool availability, rather than merely a harder version of the same accuracy evaluation.

\FloatBarrier

\section{Additional Trace-Level Failure Analysis}
\label{sec:results-failurecases}

This appendix provides the failure taxonomy, marginal failure distributions, and failure co-occurrence heatmaps used for the trace-level diagnostics summarized in Section~\ref{sec:main-failure}.
The taxonomy is non-exclusive: multiple labels may be assigned to a single failed trace.
Marginal distributions show which failure types appear most frequently for each model, while co-occurrence heatmaps show how failure modes are coupled under different tool-catalog conditions.

\begin{table}[!h]
\centering
\caption{Failure-type taxonomy used in human trace analysis.}
\label{tab:failure-taxonomy}
\small
\setlength{\tabcolsep}{6pt}
\renewcommand{\arraystretch}{1.1}
\begin{tabularx}{\columnwidth}{>{\raggedright\arraybackslash}p{0.28\columnwidth} X}
\toprule
\textbf{Label} & \textbf{Definition} \\
\midrule
\textbf{Plan Error (\(e_1\))} &
The initial plan is ineffective (logically incorrect, or missing essential steps/subgoals). \\

\textbf{Tool Selection Error (\(e_2\))} &
At least one call selects an irrelevant tool or fails to use a suitable tool for a subproblem. \\

\textbf{Tool Hallucination (\(e_3\))} &
At least one call references a tool name that does not exist in the provided tool list. \\

\textbf{Wrong Parameter Value (\(e_4\))} &
At least one call uses schema-valid argument \emph{types} but incorrect/irrelevant values (e.g., wrong constants/inputs, mismatched intermediate values). \\

\textbf{Formatting Error (\(e_5\))} &
At least one call violates the input schema or fails JSON conversion (e.g., wrong types/missing fields, malformed JSON, tool calls in plain text). \\

\textbf{Thought Error (\(e_6\))} &
Reasoning is incorrect given the final answer/tool logic, contradicts observations, or hallucinates \texttt{Action}/\texttt{Observation} to bypass interaction. \\

\textbf{Observation Omission (\(e_7\))} &
Two or more consecutive \texttt{Thought} steps without incorporating an intervening tool \texttt{Observation} when a tool interaction is expected. \\

\textbf{Repeated Call (\(e_8\))} &
The exact same \emph{(tool, arguments)} pair is called 2 or more times after being executed (re-calls with different arguments are not counted). \\

\textbf{Incomplete Execution (\(e_9\))} &
No final answer is produced (e.g., timeout, loops/excessive repeats, or exceeding the maximum turn budget). \\
\bottomrule
\end{tabularx}
\vspace{-2mm}
\end{table}

\begin{figure}[!h]
\centering
\includegraphics[width=0.9\columnwidth]{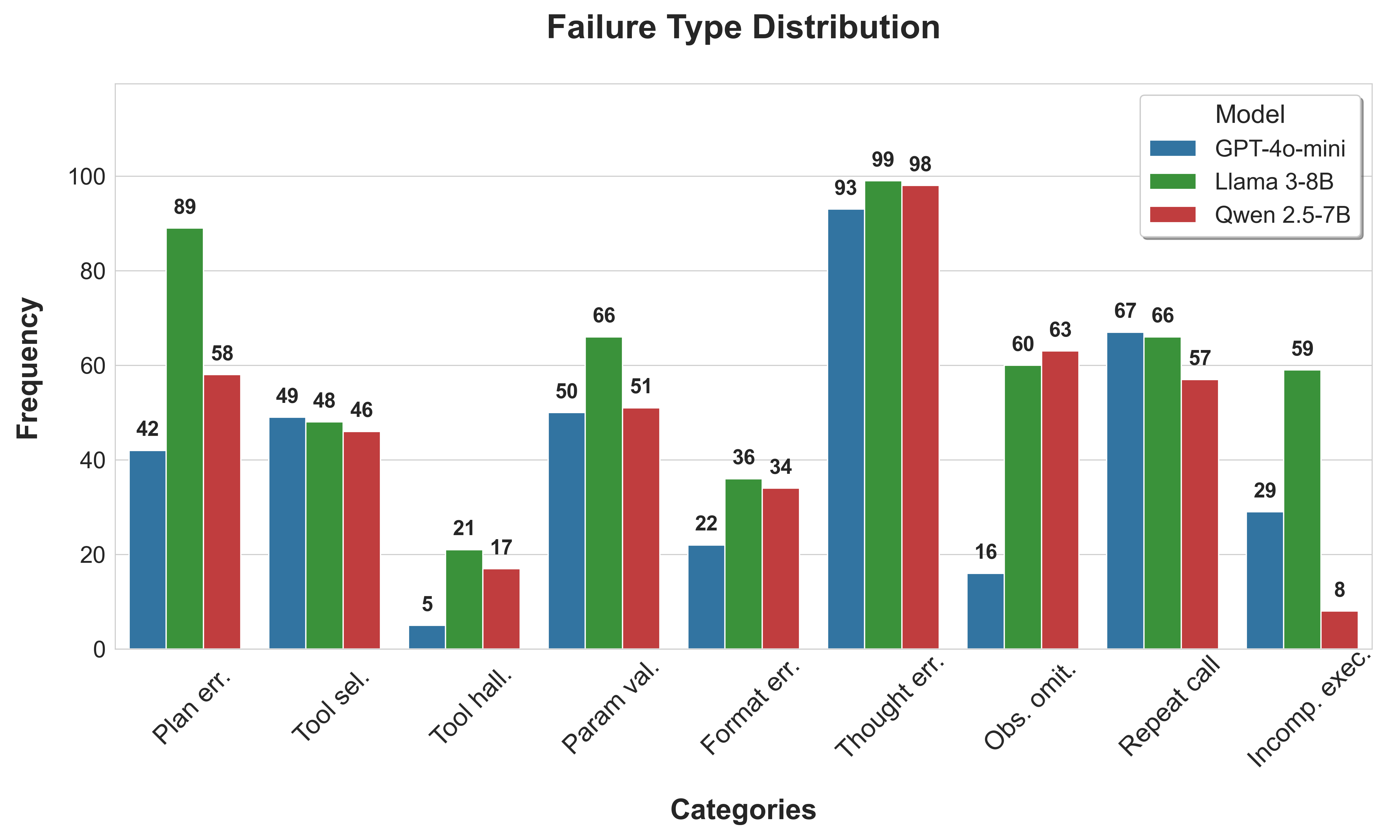}
\caption{\textbf{Failure cases under Gold-present (Level~3, \(k=5\)).}
Counts are computed from 100 failed instances per model; multiple labels per instance are allowed.}
\label{fig:app-failure-distribution}
\end{figure}

\FloatBarrier

\paragraph{Failure co-occurrence.}
To analyze coupled failure patterns, we compute error co-occurrence heatmaps.
Let \(e_a\) and \(e_b\) denote failure labels.
For each model and tool-catalog condition, we compute
\[
H_{a,b}=P(e_a\mid e_b),
\]
the fraction of failed traces containing error \(e_b\) that also contain error \(e_a\).
These conditional statistics summarize which failure modes tend to appear together under each setting and should be interpreted as diagnostic associations rather than causal estimates.

\begin{figure}[!h]
    \centering
    \includegraphics[width=0.95\textwidth]{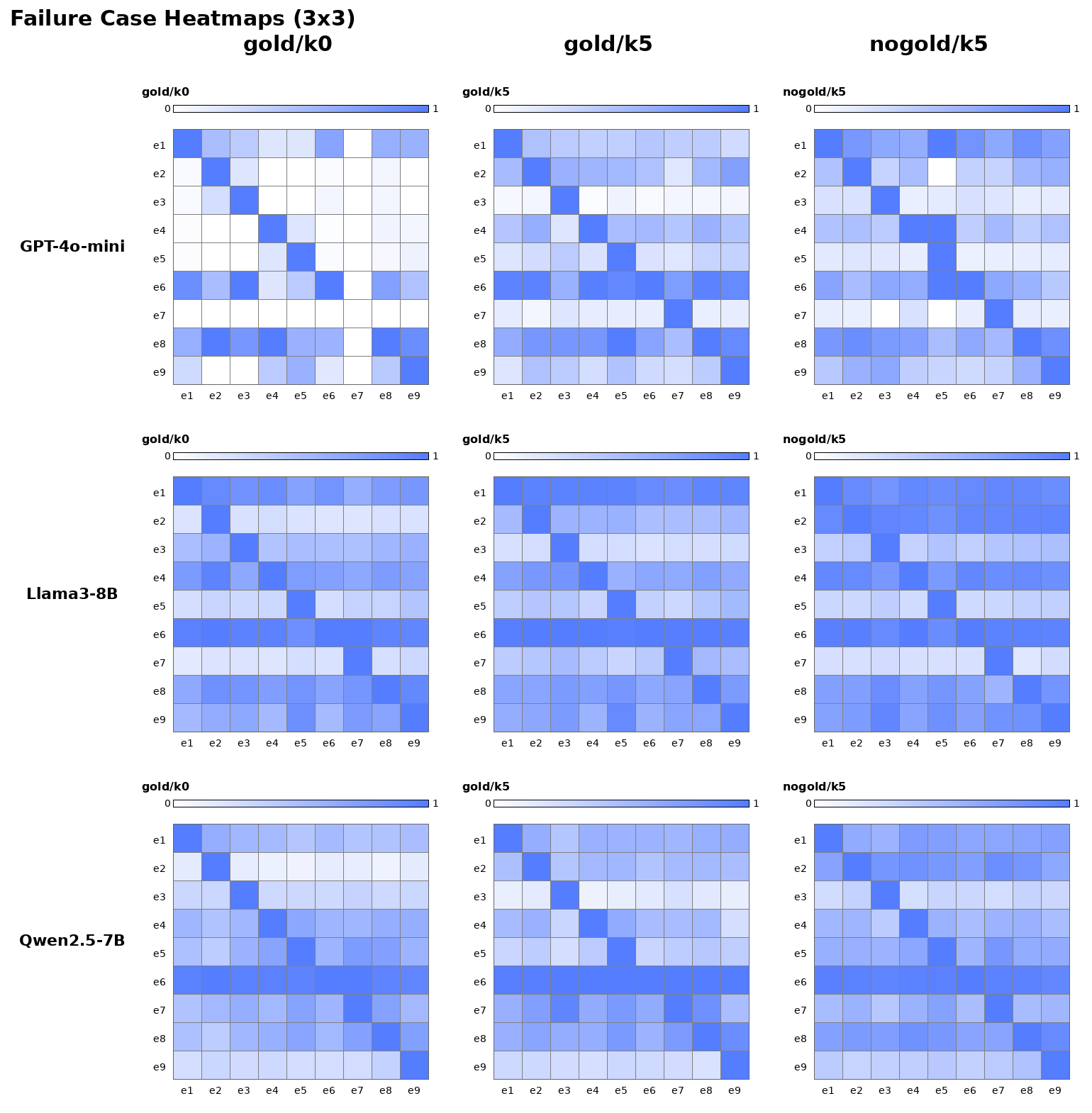}
    \caption{
    \textbf{Failure co-occurrence heatmaps across models and tool-catalog conditions.}
    Columns correspond to the Gold-only, Gold-present, and Distractors-only conditions.
    For Gold-present and Distractors-only, we use a distractor budget of \(k=5\) with Level~3 sampling.
    Rows correspond to GPT-4o-mini, Llama 3-8B, and Qwen 2.5-7B.
    Each cell reports \(P(e_a \mid e_b)\) over failed traces.
    The labels \(e_1\)--\(e_9\) correspond to the failure taxonomy listed in Table~\ref{tab:failure-taxonomy}.
    }
    \label{fig:app-failure-heatmaps}
\end{figure}

Figure~\ref{fig:app-failure-heatmaps} shows that tool-catalog conditions change not only accuracy, but also the structure of model failures.
For Llama 3-8B, Tool Selection Error is relatively localized in the Gold-only condition, becomes more broadly coupled with other errors under the Gold-present condition, and is strongly coupled with nearly all other errors under the Distractors-only condition.
This suggests that removing gold tools changes tool selection from an isolated mistake into a failure mode that co-occurs with broader execution breakdown.

The same heatmaps reveal model-specific differences under the same catalog condition.
In the Gold-only condition, Planning Error is selectively coupled with downstream errors for GPT-4o-mini, more uniformly coupled for Llama 3-8B, and moderately coupled for Qwen 2.5-7B.
Thus, \ToolMATH exposes differences in failure organization, not merely differences in final answer accuracy.

\FloatBarrier

\section{Representative correct traces in the \emph{Distractors-only} condition}
\label{app:distractor_only_success}

This appendix presents representative episodes that are solved correctly under the
\emph{Distractors-only} condition (gold tools removed), where the model executes one or more
available distractor tools as part of its solution trace.
Our goal is to qualitatively illustrate the kinds of tool executions and compositions that
can arise even in tool-insufficient environments.

\paragraph{Selection criterion.}
We collect episodes that satisfy:
(i) evaluation is conducted under the \emph{Distractors-only} tool list,
(ii) the normalized final answer is correct, and
(iii) the trace contains at least one executed tool call (a parsed \texttt{Action} followed by an \texttt{Observation}).

\paragraph{Example case for GPT-4o-mini.}
\begin{promptbox}
Problem:
How many vertical asymptotes does the graph of
$y=\dfrac{1}{x^3-3x^2+3x-1}$ have?

Tools (distractor-only;):
- solve_rational_equation_quadratic
    -"description": "Solve equation a x/(x-b) + (c x^2 + d)/x = k. Returns real solutions excluding poles."
    -"inputs": {
          "a": "float",
          "b": "float",
          "c": "float",
          "d": "float",
          "k": "float"
        }
- line_coefficients_from_point_direction
    ...
- angle_from_three_points
    ...
- sum_first_digits_of_fraction_sum
    ...
- least_multiple_with_num_divisors
    ...
- max_gcd_minus_a
    ...
- t_for_pairwise_sums_cubic
    ...
- real_part_from_modulus_and_imag
    ...
- evaluate_quadratic
    ...
- c_from_floor_and_fractional_roots
    ...

Plan:
1. Vertical asymptotes occur at real $x$ values where the denominator $x^3-3x^2+3x-1$ equals $0$.
2. To find the real roots of the denominator, use the tool `[TOOL:solve_rational_equation_quadratic]`, which numerically returns real solutions of a cubic polynomial determined by its parameters.
3. Choose parameters so the tool's internal cubic matches $x^3-3x^2+3x-1=0$.
4. Count the number of distinct real roots returned.

Thought: Find real zeros of the denominator via a distractor tool and count distinct roots.
Action: {"name":"solve_rational_equation_quadratic","arguments":{"a":-8.0,"b":-1.0,"c":1.0,"d":-1.0,"k":-4.0}}
Observation (solve_rational_equation_quadratic, {"a":-8.0,"b":-1.0,"c":1.0,"d":-1.0,"k":-4.0}): [np.float64(1.0)]
Thought: The distinct real root set is $\{1\}$, so there is exactly one vertical asymptote.
ANSWER: 1
\end{promptbox}

\clearpage

\paragraph{Example case for Qwen2.5-7B.}
\begin{promptbox}
Problem:
What is $2a+3b$, where $a=2-i$ and $b=-1+i$?

Tools (distractor-only; names only):
- add_vectors_2d
    - "description": "Add two 2D vectors."
    - "inputs": { 
        "v1": "list[float]", 
        "v2": "list[float]" 
        }
- line_coefficients_from_point_direction
    ...
- monic_quadratic_from_surd_shift
    ...
- count_isosceles_from_lengths
    ...
- solve_c_for_polynomial_value
    ...
- add_polynomials
    ...
- line_from_point_normal_2d
    ...
- simplify_nested_radicals
    ...
- sum_coordinates_intersections_parabolas
    ...
- centroid_distance_from_vertex
    ...

Plan:
1. Rewrite each complex number as a 2D vector of its real and imaginary parts:
   $a \mapsto [2,-1]$ and $b \mapsto [-1,1]$.
2. Compute the scaled vectors by repeated addition:
   - $2a = a+a$ using `[TOOL:add_vectors_2d]`.
   - $3b = b+b+b$ using `[TOOL:add_vectors_2d]` twice.
3. Add the results to obtain $2a+3b$ as a 2D vector $[\,\Re,\,\Im\,]$.
4. Convert the final 2D vector back to complex form $\Re+\Im i$.

Thought: Represent $a,b$ as 2D vectors and use repeated vector-addition to form $2a+3b$.
Action: {"name":"add_vectors_2d","arguments":{"v1":[2,-1],"v2":[2,-1]}}
Observation (add_vectors_2d, {"v1":[2,-1],"v2":[2,-1]}): [4,-2]
Thought: Now, let's build $3b$ by adding $b$ twice more starting from $b+b$.
Action: {"name":"add_vectors_2d","arguments":{"v1":[-1,1],"v2":[-1,1]}}
Observation (add_vectors_2d, {"v1":[-1,1],"v2":[-1,1]}): [-2,2]
Thought: Let's add $2b$ and $b$ to get $3b$.
Action: {"name":"add_vectors_2d","arguments":{"v1":[-2,2],"v2":[-1,1]}}
Observation (add_vectors_2d, {"v1":[-2,2],"v2":[-1,1]}): [-3,3]
Thought: We can now add $2a$ and $3b$.
Action: {"name":"add_vectors_2d","arguments":{"v1":[4,-2],"v2":[-3,3]}}
Observation (add_vectors_2d, {"v1":[4,-2],"v2":[-3,3]}): [1,1]
Thought: We have already obtained the result vector.
ANSWER: 1+1j
\end{promptbox}

\section{Source dataset and license}
\label{app:source_dataset_license}

\ToolMATH is derived from the Mathematics Aptitude Test of Heuristics (MATH) dataset~\citep{hendrycks2021math}.
We use the Hugging Face distribution \texttt{qwedsacf/competition\_math}, which contains competition mathematics problems and step-by-step solutions from sources such as AMC 10, AMC 12, and AIME.
Each example includes a problem statement, a worked solution, a difficulty level, and a problem type.
The Hugging Face dataset card links its licensing information to the original Hendrycks MATH repository, whose license is the MIT License.
We cite the original MATH paper and use the dataset only as the source corpus for constructing derived tool-grounded benchmark instances.


\section{Impact Statement} \ToolMATH is a benchmark for evaluating tool-augmented language models under controlled multi-tool conditions. It aims to improve the reliability and interpretability of tool-use evaluation by using correctness-checkable tasks, validating tool specifications, and stress-testing models under tool-list redundancy and long-horizon execution. This can help surface brittle behaviors and support safer deployment of tool-augmented systems.

\clearpage
\section*{NeurIPS Paper Checklist}

\begin{enumerate}

\item {\bf Claims}
    \item[] Question: Do the main claims made in the abstract and introduction accurately reflect the paper's contributions and scope?
    \item[] Answer: \answerYes{}
    \item[] Justification: The abstract and introduction describe \ToolMATH as a controlled diagnostic benchmark for long-horizon tool use, emphasizing systematic distractor construction, behavior-conditioned metrics, and trace-level failure diagnostics. These claims are supported by the dataset construction, evaluation setup, results, and limitations sections.

\item {\bf Limitations}
    \item[] Question: Does the paper discuss the limitations of the work performed by the authors?
    \item[] Answer: \answerYes{}
    \item[] Justification: The paper includes a dedicated Limitations section discussing the math-grounded scope, dependence on prompting and controller details, and the diagnostic rather than causal interpretation of trace-level co-occurrence statistics.

\item {\bf Theory assumptions and proofs}
    \item[] Question: For each theoretical result, does the paper provide the full set of assumptions and a complete (and correct) proof?
     \item[] Answer: \answerNA{}
    \item[] Justification: The paper does not present theoretical results, theorems, or formal proofs. The mathematical notations are only used to define dataset construction, tool-catalog conditions, and evaluation metrics.

    \item {\bf Experimental result reproducibility}
    \item[] Question: Does the paper fully disclose all the information needed to reproduce the main experimental results of the paper to the extent that it affects the main claims and/or conclusions of the paper (regardless of whether the code and data are provided or not)?
    \item[] Answer: \answerYes{}
    \item[] Justification: The paper describes the benchmark construction pipeline, tool-wise and question-wise validation, distractor sampling procedure, tool-use protocols, decoding settings, stopping criteria, and evaluation metrics in the main paper and appendix. These details provide the information needed to reproduce or verify the main experimental claims, subject to access to the evaluated proprietary models.

\item {\bf Open access to data and code}
    \item[] Question: Does the paper provide open access to the data and code, with sufficient instructions to faithfully reproduce the main experimental results, as described in supplemental material?
    \item[] Answer: \answerYes{}
    \item[] Justification: The paper includes a public release of the full benchmark artifacts and reproduction code.

\item {\bf Experimental setting/details}
    \item[] Question: Does the paper specify all the training and test details (e.g., data splits, hyperparameters, how they were chosen, type of optimizer) necessary to understand the results?
    \item[] Answer: \answerYes{}
    \item[] Justification: The paper specifies the evaluated tool-catalog conditions, distractor levels and budgets, benchmark splits, models, frameworks, decoding temperatures, timeouts, and answer normalization at the level needed to interpret the reported results. Since the paper evaluates existing models rather than training new ones, optimizer and training hyperparameters are not applicable.

\item {\bf Experiment statistical significance}
    \item[] Question: Does the paper report error bars suitably and correctly defined or other appropriate information about the statistical significance of the experiments?
    \item[] Answer: \answerNo{}
    \item[] Justification: The current draft reports aggregate accuracies and behavior-conditioned metrics but does not include confidence intervals, error bars, or significance tests. The results should therefore be interpreted as descriptive comparisons under fixed benchmark settings rather than statistical significance claims.

\item {\bf Experiments compute resources}
    \item[] Question: For each experiment, does the paper provide sufficient information on the computer resources (type of compute workers, memory, time of execution) needed to reproduce the experiments?
    \item[] Answer: \answerYes{}
    \item[] Justification: The appendix reports computer resources such as per-question and per-tool timeouts, and wall-clock runtime.
    
\item {\bf Code of ethics}
    \item[] Question: Does the research conducted in the paper conform, in every respect, with the NeurIPS Code of Ethics \url{https://neurips.cc/public/EthicsGuidelines}?
    \item[] Answer: \answerYes{}
    \item[] Justification: The research conducted in the paper conformed, in every respect, with the NeurIPS Code of Ethics.

\item {\bf Broader impacts}
    \item[] Question: Does the paper discuss both potential positive societal impacts and negative societal impacts of the work performed?
    \item[] Answer: \answerYes{}
    \item[] Justification: We discuss potential societal impacts in the appendix.
    
\item {\bf Safeguards}
    \item[] Question: Does the paper describe safeguards that have been put in place for responsible release of data or models that have a high risk for misuse (e.g., pre-trained language models, image generators, or scraped datasets)?
    \item[] Answer: \answerNA{}
    \item[] Justification: The paper does not release a high-risk pretrained model, image generator, or scraped dataset, etc. The benchmark is math-grounded and does not contain personal or sensitive data, so special release safeguards of this type are not applicable.

\item {\bf Licenses for existing assets}
    \item[] Question: Are the creators or original owners of assets (e.g., code, data, models), used in the paper, properly credited and are the license and terms of use explicitly mentioned and properly respected?
    \item[] Answer: \answerYes{}
    \item[] Justification: The paper cites the MATH dataset and the evaluated model families, and it explicitly states the licenses or terms of use for the MATH dataset in the appendix.

\item {\bf New assets}
    \item[] Question: Are new assets introduced in the paper well documented and is the documentation provided alongside the assets?
    \item[] Answer: \answerYes{}
    \item[] Justification: New assets introduced in the paper are well documented in the main text and appendix. The accompanying documentations in the dataset and code include the same schema, validation details, and usage instructions.

\item {\bf Crowdsourcing and research with human subjects}
    \item[] Question: For crowdsourcing experiments and research with human subjects, does the paper include the full text of instructions given to participants and screenshots, if applicable, as well as details about compensation (if any)? 
    \item[] Answer: \answerNA{}
    \item[] Justification: The paper does not report crowdsourcing experiments or research with human subjects.

\item {\bf Institutional review board (IRB) approvals or equivalent for research with human subjects}
    \item[] Question: Does the paper describe potential risks incurred by study participants, whether such risks were disclosed to the subjects, and whether Institutional Review Board (IRB) approvals (or an equivalent approval/review based on the requirements of your country or institution) were obtained?
    \item[] Answer: \answerNA{}
    \item[] Justification: The paper does not involve human-subject experiments or collection of personal data from participants, so IRB approval or equivalent human-subject review is not applicable.

\item {\bf Declaration of LLM usage}
    \item[] Question: Does the paper describe the usage of LLMs if it is an important, original, or non-standard component of the core methods in this research? Note that if the LLM is used only for writing, editing, or formatting purposes and does \emph{not} impact the core methodology, scientific rigor, or originality of the research, declaration is not required.
    \item[] Answer: \answerYes{}
    \item[] Justification: LLMs are used as part of the benchmark construction and validation pipeline, including tool construction, validation, and model evaluation.

\end{enumerate}

\end{document}